\documentclass[manuscript,screen,nonacm]{acmart}

\usepackage{graphicx}
\usepackage{multirow}
\usepackage{amsmath,amsfonts}
\usepackage{amsthm}
\usepackage{mathrsfs}
\usepackage[title]{appendix}
\usepackage{xcolor}
\usepackage{textcomp}
\usepackage{manyfoot}
\usepackage{booktabs}
\usepackage{algorithm}
\usepackage{algorithmicx}
\usepackage{algpseudocode}
\usepackage{listings}
\usepackage{csquotes}
\usepackage{tabularx}
\usepackage{listings}
\usepackage{xcolor}
\usepackage{booktabs}
\usepackage{makecell}
\usepackage{pifont}
\usepackage{array}

\newcommand{\cmark}{\ding{51}}
\newcommand{\xmark}{\ding{55}}

\lstdefinestyle{promptstyle}{
    basicstyle=\ttfamily\scriptsize,
    breaklines=true, 
    breakatwhitespace=true,
    frame=single,
    rulecolor=\color{black!30},
    backgroundcolor=\color{gray!5},
    columns=fullflexible,
    keepspaces=true,
    captionpos=b,
    extendedchars=true,
    showstringspaces=false,
    tabsize=2
}

\setcopyright{acmlicensed}
\copyrightyear{2024}
\acmYear{2024}
\acmDOI{XXXXXXX.XXXXXXX}

\AtBeginDocument{%
  }

\begin{document}

\title[CMASE]{Computational Multi-Agent Society Experiments: Social Modeling Framework Based on Generative Agents}

\author{Hanzhong Zhang}
\email{armihiabelliard@gmail.com}
\affiliation{%
  \institution{University of Exeter}
  \city{Exeter}
  \country{UK}
}

\author{Muhua Huang}
\affiliation{%
  \institution{Stanford University}
  \city{Stanford}
  \state{CA}
  \country{USA}}
\email{muhua@stanford.edu}

\author{Jindong Wang}
\authornote{Corresponding author.}
\email{jwang80@wm.edu}
\affiliation{%
  \institution{William \& Mary}
  \city{Williamsburg}
  \state{VA}
  \country{USA}
}

\begin{CCSXML}
<ccs2012>
   <concept>
       <concept_id>10003120.10003121.10003122</concept_id>
       <concept_desc>Human-centered computing~HCI design and evaluation methods</concept_desc>
       <concept_significance>500</concept_significance>
       </concept>
   <concept>
       <concept_id>10003120.10003130</concept_id>
       <concept_desc>Human-centered computing~Collaborative and social computing</concept_desc>
       <concept_significance>500</concept_significance>
       </concept>
 </ccs2012>
\end{CCSXML}

\ccsdesc[500]{Human-centered computing~Collaborative and social computing}
\ccsdesc[500]{Human-centered computing~HCI design and evaluation methods}

\renewcommand{\shortauthors}{Zhang et al.}

\begin{abstract}
Computational social science has extensively developed descriptive, interpretative, and predictive modeling, but comprehensive modeling, which both admits human intervention into a simulated society and aims to forecast outcomes, remains underexplored. Existing generative agent-based platforms scale up populations or reproduce complex social scenarios, but researcher involvement is generally implemented through initial configuration, external runtime controls, or post-hoc analysis rather than as an explicit in-world role within the ongoing simulation. This paper introduces Computational Multi-Agent Society Experiments (CMASE), a generative agent-based framework that explicitly integrates the researcher into simulation runtime through role-based, in-world participation and traceable intervention channels. By treating the simulation as a simulated ethnographic field, CMASE shifts the researcher from an external operator to an embedded participant, and is designed to achieve three core capabilities: (1) enabling real-time human-computer interaction that allows researchers to dynamically embed themselves into the system to characterize complex social intervention processes; (2) combining controlled computational experimentation with ethnography-inspired observation, interviewing, and interpretation within the simulated field; and (3) supporting forward exploration of intervention trajectories and their downstream social responses within the simulation. Methodologically, CMASE realizes these capabilities through agents whose perception, memory, and emotion are driven by a large language model, a composable environment built from a texture and area map, and a typology of researcher roles spanning observation, interviewing, embedded participation, and intervention, so that an intervention enters the society as a social event rather than an exogenous parameter. We evaluate the framework through computational replications of two published field studies from different disciplines, one on urban greenery and social fragmentation and one on cooperation dynamics in a public goods game, using two substantially different language models. CMASE reproduces reported statistical patterns and yields process-level records that support mechanism-oriented interpretation within the simulated settings, while a controlled comparison shows that an intervention delivered as an embedded social event leaves an observable, auditable social process that a parametric manipulation does not. These findings demonstrate CMASE's methodological value for studying intervention processes in small- and medium-scale social settings, highlighting its potential to advance interdisciplinary integration in the social sciences. The official code is available at: \url{https://github.com/armihia/CMASE}.

\end{abstract}

\keywords{LLM-based Agents, Society Simulation, Generative Agent-Based Modeling, Computational Social Science}

\maketitle

\section{Introduction}
\label{sec1}

Social science explores human behaviors and social structures to understand how society operates. A difficult methodological question in this field concerns \emph{intervention processes}: how a social system evolves before, during, and after a researcher or policymaker acts upon it. Traditional sociological research addresses such questions through instruments that rely on human participants, such as survey questionnaires and laboratory or field experiments \cite{groves2011survey,falk2009lab}.

These methods are commonly used to test theoretical hypotheses, understand social phenomena, and predict collective outcomes, and they yield highly realistic data. However, they capture outcomes only at discrete measurement points, are costly and difficult to scale, and carry ethical risks when the intervention under study may harm participants \cite{lazer2009computational}. As a result, the process of an intervention, that is, how it is perceived, resisted, or amplified by the individuals it targets, remains largely unobservable with these instruments. The goal of this paper is to provide a computational methodology that makes intervention processes observable, manipulable, and interpretable within a simulated society.

To reduce cost and scale constraints, researchers have long turned to agent-based modeling (ABM), which simulates societies as populations of rule-following agents \cite{Bonabeau2002, macy2002factors}. Yet in ABM, agent behavior is dictated by mathematical formulas or hand-crafted rules whose parameters are specified by the modeler, so simulated dynamics inevitably reflect the modeler's prior mental model rather than human-like decision-making. Generative agent-based modeling (GABM) overcomes this limitation by delegating each agent's decisions to a large language model (LLM): instead of imposing explicit behavioral rules, the modeler specifies only each agent's persona (e.g., age, gender, temperament, occupation) and environmental constraints, and the LLM, drawing on human behavioral patterns embedded in its training data, generates context-dependent actions \cite{vezhnevets2023generative, ghaffarzadegan2024generative, park2023generative}. This property is exactly what our task requires: if intervention processes are to be studied in simulation, agents must respond to interventions in ways that are human-like and spontaneous rather than scripted. Building on this paradigm, LLM-based multi-agent platforms such as AgentSociety \cite{piao2025agentsociety}, OASIS \cite{yang2024oasis}, and GenSim \cite{tang2024gensim} scale generative simulations to tens of thousands or even a million agents.

Despite this progress, existing platforms share a structural limitation that can be made precise using the taxonomy of Xue et al. \cite{xue2024computational}, which classifies sociological modeling by (1) whether human intervention is involved and (2) whether the goal is feature explanation or outcome forecasting, yielding descriptive, interpretative, predictive, and comprehensive modeling. The first three types are well developed; \emph{comprehensive modeling}, which admits human intervention into the system \emph{and} aims to forecast outcomes, remains relatively unexplored. Comprehensive modeling imposes a requirement that none of the platforms above satisfies: the intervention process itself must be endogenously modeled, capturing the system's state before the intervention, tracing its evolution during the intervention, and producing forecasts of its consequences afterward, all within one continuous run. Current GABM platforms, by contrast, generally keep the researcher outside the simulated society as a designer, operator, or post-hoc analyst. Although agent behavior can be scripted and some platforms provide runtime controls, researcher participation itself is not explicitly represented as an in-world role through which the researcher can observe unfolding interactions, adapt subsequent actions, and remain socially attributable to the agents being studied. This is the critical gap addressed in this paper: the absence of a framework that operationalizes the researcher as a runtime participant in generative social simulation.

In this paper, we propose \textbf{C}omputational \textbf{M}ulti-\textbf{A}gent \textbf{S}ociety \textbf{E}xperiments (\textbf{CMASE}), which is, to the best of our knowledge, the first GABM framework to support real-time, human-in-the-loop intervention during simulation runtime, thereby providing an operational path toward comprehensive modeling. Specifically, CMASE treats the simulation as a simulated ethnographic field: the researcher is no longer an external operator of a static model but an embedded participant who can observe agents, interview them, act alongside them as a peer under the same behavioral constraints, or reshape the environment through events at any moment during a run. Crucially, these interventions are not silent parameter overwrites; they enter agents' perception, memory, and emotional appraisal, so that the intervention itself becomes a modeled social variable whose process-level effects can be traced. 

We position our contribution as primarily methodological: CMASE contributes (i) an architecture that explicitly represents researcher participation at runtime through role-dependent permissions, allowing researcher actions to enter agents' ordinary perception, memory, and interaction processes as attributable social events; (ii) an ethnography-inspired protocol in which the researcher can observe, interview, participate, and intervene within an unfolding simulated field, including adapting subsequent actions in response to ongoing interactions while preserving a traceable event record; and (iii) empirical evaluation through two computational replications of published field studies, examining whether CMASE can reproduce reported social patterns while supporting additional process-level inquiry within the simulated settings. The first reproduces the reported quantitative relationship between urban greenery and perceived social fragmentation and uses in-world interviews and participant observation to examine candidate process accounts within the simulation. The second reproduces reported cooperation dynamics in a repeated public goods game and then uses the reconstructed setting to test the intervention typology under controlled conditions. Across the two studies, CMASE is instantiated on both a frontier commercial model and a substantially smaller open-weights model, demonstrating that the framework can support these forms of inquiry across different generative substrates. Accordingly, we scope our claims to small- and medium-scale social settings in which process-level intervention and interpretive understanding are central (e.g., community dynamics, organizational micro-processes, and policy pre-testing), rather than to the social sciences in general. In summary, CMASE is designed to achieve three research objectives:
\begin{itemize}
\item To enable real-time human--computer interaction (HCI) within social simulations, replacing the static, predefined configuration of existing platforms with a dynamic environment in which researchers are embedded directly in the simulation loop, continuously observing and adjusting complex social processes as they unfold.
\item To implement and evaluate researcher-led interventions through an ethnography-inspired approach, allowing researchers to test how specific changes in variables or agent interactions influence the generative logic of social phenomena, rather than treating simulation as a black box.
\item To support exploration of intervention trajectories, helping researchers examine how simulated social processes develop after an intervention and generate process-level hypotheses for validation with human evidence.
\end{itemize}

\section{Related Work}\label{sec2}

\subsection{Simulating Social Behavior with Large Language Models}\label{subsec2.1}

Social simulation has been an important tool for understanding and analyzing social phenomena. Early studies mainly relied on rule-based artificial agent models, namely agent-based modeling (ABM)\cite{Bonabeau2002, macy2002factors}. Traditional ABMs often depend on manually crafted heuristic rules, making it difficult to approximate the complex features of real human cognition, language, and emotion \cite{ma2024computational, wu2023smart, jager2022can}. In recent years, with the capabilities of large language models (LLMs) in simulating human behavior, research on building social simulators based on LLMs has rapidly advanced \cite{park2023generative}. GABM has gradually become one of the core methods for simulating complex social behavior \cite{vezhnevets2023generative}. In this individual-level approach, mechanistic models of agent interaction are combined with LLMs, and each agent makes decisions after communicating with the LLM. This method minimizes the need for modelers to make assumptions about human decision-making, instead leveraging the vast data embedded in LLMs to capture human behaviors and choices \cite{ghaffarzadegan2024generative}.

Existing GABM systems can be roughly categorized into two main research directions. The first line of research focuses on realistic simulations of specific social scenarios, aiming to foster competition or collaboration among agents to generate social insights or support policy design. Li et al. \cite{li2024agent} developed Agent Hospital, where LLM-driven doctor agents interact with large numbers of patients to evolve transferable medical decision-making capabilities. Yu et al. \cite{yu2024fincon} designed FinCon, a hierarchical agent system modeled after financial institutions, using language interaction and belief reinforcement to enhance collective collaboration in complex investment tasks. TrendSim \cite{zhang2024trendsim} focuses on topic propagation on social media and its vulnerability to poisoning attacks, providing intervention strategies for platform governance. ElectionSim \cite{zhang2024electionsim} reconstructs election scenarios using millions of social media records, achieving high-fidelity voter modeling and policy preference prediction. Zhao et al. \cite{zhao2023competeai} simulated competitive dynamics between restaurant agents and customer agents to align with existing market and sociological theories.

The second line of GABM research emphasizes building reusable, scalable simulation platforms that support general modeling tasks. Wang et al. \cite{wang2025yulan} divide the development of LLM-based social simulation platforms into two stages. In the first stage, researchers focus on simulating specific scenarios with typically fewer than 1,000 agents (such as \cite{zhou2025sotopia}). By proposing YuLan-OneSim, they aim to push the field into the next stage by introducing environmental construction, agent evolution, and feedback integration mechanisms, thus enabling simulations of larger and more complex social processes. Frameworks of similar scale now support simulations with over 10,000 agents. AgentSociety \cite{piao2025agentsociety} recreates contexts to evaluate policy interventions and the emergence of social mechanisms. AgentScope \cite{gao2024agentscope} offers modular design, no-code interfaces, and distributed deployment, significantly lowering the barrier to building multi-agent systems. Socioverse \cite{zhang2025socioverse} integrates data from tens of millions of real users to provide a reliable foundation for modeling multiple subdomains such as politics and economics. GenSim \cite{tang2024gensim} supports up to 100,000 agents and includes error correction mechanisms, emphasizing generalizability and long-term robustness. OASIS \cite{yang2024oasis} uses open social platforms for simulation, supporting dynamic networks, recommendation systems, and multi-behavior modeling.

However, exponential increases in agent quantity alone are not sufficient to effectively capture the essence of social complexity. Recent research on freely formed AI collectives similarly suggests that social diversity, innovation, and self-regulation emerge less from scale than from the dynamics of open-ended interaction \cite{lai2024}. In particular, without mechanisms for understanding social context and interactive feedback, large-scale simulations often struggle to offer human-interpretable responses to questions about how social structures and norms form or institutions evolve.

Unlike early studies that relied on extensive handcrafted heuristic rules to dictate specific actions, our approach only establishes a priori environmental constraints and cognitive workflows. Within these boundaries, the system relies entirely on the LLM-based agents' capacity to replicate human behavior to drive the simulation. This ensures that the resulting social dynamics are systematic, emergent properties rather than cherry-picked, pre-defined outcomes.

\subsection{Human-AI Collaborative Modeling in Social Systems}\label{subsec2.2}

Computational social science explores social structures, behavioral mechanisms, and institutional evolution in simulated computational environments through formalized approaches \cite{lazer2009computational}. Within this field, the computational experiments paradigm aims to systematically manipulate variables in virtual societies, thereby extending the boundaries of traditional experimentation \cite{xue2023computational}. In this context, LLM-based GABM is not only a representational tool for social processes but also holds the potential to become a platform for computational experimentation. Sato \cite{sato2024sociological} highlights the importance of meaning and reflexivity in bridging the gap between agent-based modeling and social theory. This implies that for agent-based modeling to contribute to theoretical advancement in the social sciences, it must go beyond scale amplification and incorporate workflows that offer human interpretability. Therefore, beyond the expansion of scale, a more socially grounded, interactive, and reflexive human–AI sociological framework is needed, in which agents are no longer passive generators of behavior but are embedded within social networks and actively participate in meaning-making processes \cite{tsvetkova2024new}.

Recent research in this direction has focused on the subfield of human-AI co-creation. Broadly speaking, GABM belongs to this category: through human-designed environments, agent adaptation, and human observational feedback, it enables collaborative modeling between humans and AI. Some studies emphasize interactive environment co-construction. For example, HABITAT 3.0 \cite{puig2023habitat} supports real-time task collaboration between humans and humanoid agents in virtual spaces. As this field has progressed, researchers have begun to explore more creative and socially complex scenarios of human-AI co-creation, emphasizing how humans guide emergent AI behavior. The Cognitio Emergens framework \cite{lin2025cognitio} redefines human-AI collaboration as a process of mutual evolution and offers a conceptual toolkit to preserve the significance of human participation. The Pecan method \cite{lou2023pecan} enhances zero-shot collaboration between AI and humans via strategic integration and context-aware mechanisms. The Collaborative Gym framework \cite{shao2024collaborative} highlights the human role in environmental control and structural intervention during task execution, supporting asynchronous three-way interaction between humans, agents, and task environments.

A complementary methodological perspective comes from recent work at the intersection of AI and anthropology \cite{deseta2024synthetic,witteborn2026context}. Emerging discussions of artificial fieldwork consider generative virtual worlds not only as environments to be observed, but as fields in which researchers interact with virtual interlocutors through interviewing and participant observation \cite{verbuyst2026artificial}. In this view, field inquiry is shaped by unfolding encounters, including cooperation, refusal, rapport, and unexpected responses, rather than by a fixed sequence of researcher actions \cite{verbuyst2026artificial,verbuyst2025refusal,verbuyst2023failure}. This perspective motivates the researcher-facing design of CMASE: the researcher remains able to observe and interpret an ongoing simulated field and to adjust subsequent inquiry or intervention within it \cite{park2022social,park2023generative,schroeder2025llmqual}.

Park et al.'s Generative Agents \cite{park2023generative} provide an important precursor by demonstrating an interactive artificial society in which agents maintain memories, plan activities, and interact socially. CMASE builds on these agent-side capabilities by adding a researcher-facing runtime layer with explicit roles and permissions. At the Participant level, the researcher acts through the same social interface as other agents and can adjust subsequent actions as interactions unfold, while these actions retain an identifiable source within the simulation record.

Against this background, the relevant distinction among existing frameworks concerns how researcher participation is represented within the running simulation. AgentSociety \cite{piao2025agentsociety} comes closest on the intervention dimension: beyond pre-simulation configuration, its experiment toolbox supports runtime state manipulation, message notification, and agent interviews and surveys. These operations, however, are issued through external control channels (a GUI and a messaging interface), and its interviews are explicitly designed so that agents respond without interrupting their ongoing actions; the researcher thus remains an operator outside the simulated society, querying agents through a side channel that leaves no trace in the social world, rather than a social actor whose presence and actions carry social meaning. Socioverse \cite{zhang2025socioverse} grounds agents in real-user data and additionally provides an in-depth interview module through which a human can query a target agent over multiple rounds; the human here acts as an out-of-world interviewer rather than a co-present member of the simulated society, and the interview is a query channel rather than a social event within it. ElectionSim \cite{zhang2024electionsim} restricts the human role to setup and post-hoc analysis. OASIS \cite{yang2024oasis} prioritizes scale and dynamic network simulation and exposes no channel for runtime human action within the simulated society. GenSim \cite{tang2024gensim} offers global intervention and agent interview interfaces that allow users to score or revise agent behaviors at runtime, yet the human operates on the simulation from outside rather than within it. For human--AI collaboration, Collaborative Gym \cite{shao2024collaborative} supports asynchronous human--agent--environment interaction for task execution, not social simulation: the human is a co-worker, not a social actor embedded in a modeled society whose interventions become objects of agent cognition. HABITAT 3.0 \cite{puig2023habitat} similarly targets embodied task collaboration.

Across these systems, CMASE operationalizes researcher participation through three linked capabilities: (1) a \emph{runtime embedding interface} with explicit researcher roles and role-dependent permissions; (2) \emph{in-world participation} through which researcher actions are perceived, remembered, and responded to as socially attributable events; and (3) \emph{process tracing} that links the intervention source to subsequent agent responses, memories, and behavior within the same run. Human control further allows the researcher to adapt inquiry or intervention to unfolding interactions, while the same role structure can also support standardized protocols when experimental control is required.

In addition, although interview-style querying is available in several platforms (e.g., AgentSociety, GenSim, and Socioverse), in all of them the interview operates as an out-of-world query channel; none supports embedded fieldwork in which the researcher is co-present with agents and interviews function as social events that enter agents' memory and emotional appraisal, supporting ethnographic analysis of the simulated society. Table~\ref{tab:framework-comparison} summarizes the comparison along these four dimensions. To the best of our knowledge, CMASE is the first GABM framework to supply all of them, thereby supporting real-time, human-in-the-loop intervention during simulation runtime.
 
\begin{table}[t]
\centering
\caption{Comparison of CMASE with representative GABM platforms and human--AI
collaboration frameworks along the three capabilities required by
comprehensive modeling and the availability of interpretive (ethnographic)
output. \cmark~= supported; (\cmark)~= partially supported; \xmark~= not
supported.}
\label{tab:framework-comparison}
\small
\begin{tabular}{lcccc}
\toprule
\textbf{Framework} &
\makecell{\textbf{Runtime human}\\\textbf{embedding}} &
\makecell{\textbf{Intervention as}\\\textbf{social variable}} &
\makecell{\textbf{Process tracing}\\\textbf{of interventions}} &
\makecell{\textbf{Interpretive}\\\textbf{output}} \\
\midrule
AgentSociety \cite{piao2025agentsociety} & (\cmark)~(external controls) & \xmark & (\cmark)~(logged outcomes) & (\cmark)~(interviews, surveys) \\
OASIS \cite{yang2024oasis} & \xmark & \xmark & (\cmark) & \xmark \\
GenSim \cite{tang2024gensim} & (\cmark)~(global intervention) & \xmark & (\cmark) & (\cmark)~(interviews) \\
Socioverse \cite{zhang2025socioverse} & (\cmark)~(interview module) & \xmark & (\cmark) & (\cmark)~(interviews) \\
Collaborative Gym \cite{shao2024collaborative} & \cmark~(task-level) & \xmark & \xmark & \xmark \\
HABITAT 3.0 \cite{puig2023habitat} & \cmark~(embodied task) & \xmark & \xmark & \xmark \\
\midrule
\textbf{CMASE (ours)} & \cmark~ & \cmark & \cmark & \cmark \\
\bottomrule
\end{tabular}
\end{table}

\section{CMASE Framework}\label{sec3}

Computational social science leverages digital technologies, large-scale data, and computational methods to study social phenomena \cite{conte2012manifesto}. As introduced in Section~\ref{sec1}, the taxonomy of Xue et al.~\cite{xue2024computational} classifies computational sociological modeling along two dimensions, namely whether human intervention is involved and whether the goal is feature explanation or outcome forecasting, yielding descriptive, interpretative, predictive, and comprehensive modeling. Two clarifications are warranted before introducing our framework. First, this taxonomy classifies research designs by their goals, not techniques by their capabilities. GABM as a technique is goal-neutral; we categorize existing GABM practice under descriptive modeling (and, when coupled with forecasting pipelines, predictive modeling) because current platforms confine human involvement to initial configuration and post-hoc analysis; no mechanism allows a researcher to intervene while the simulation runs. The descriptive and predictive categories are distinguished by their goals (explanation vs.\ forecasting) and therefore do not overlap conceptually, even when the same technique serves both. Second, comprehensive modeling is not obtained by simply adding an intervention option to predictive modeling. Its defining requirement is that the intervention process be endogenously modeled: the system's state before the intervention, its dynamic evolution during the intervention, including how agents perceive and respond to the intervener, and the resulting post-intervention forecast must all be captured within one continuous run. CMASE operationalizes this intervention-centered requirement through three corresponding mechanisms: initial-state construction via the environment maker (pre-intervention, Section~\ref{subsec3.1}); the free action mode and event chains through which researcher actions enter agents' perception, memory, and emotion (during intervention, Sections~\ref{subsec3.2}--\ref{subsec3.4}); and continued simulation and measurement of downstream agent responses after the intervention (Section~\ref{subsec3.5}).
 
\begin{figure}[h]
\centering
\includegraphics[width=0.9\textwidth]{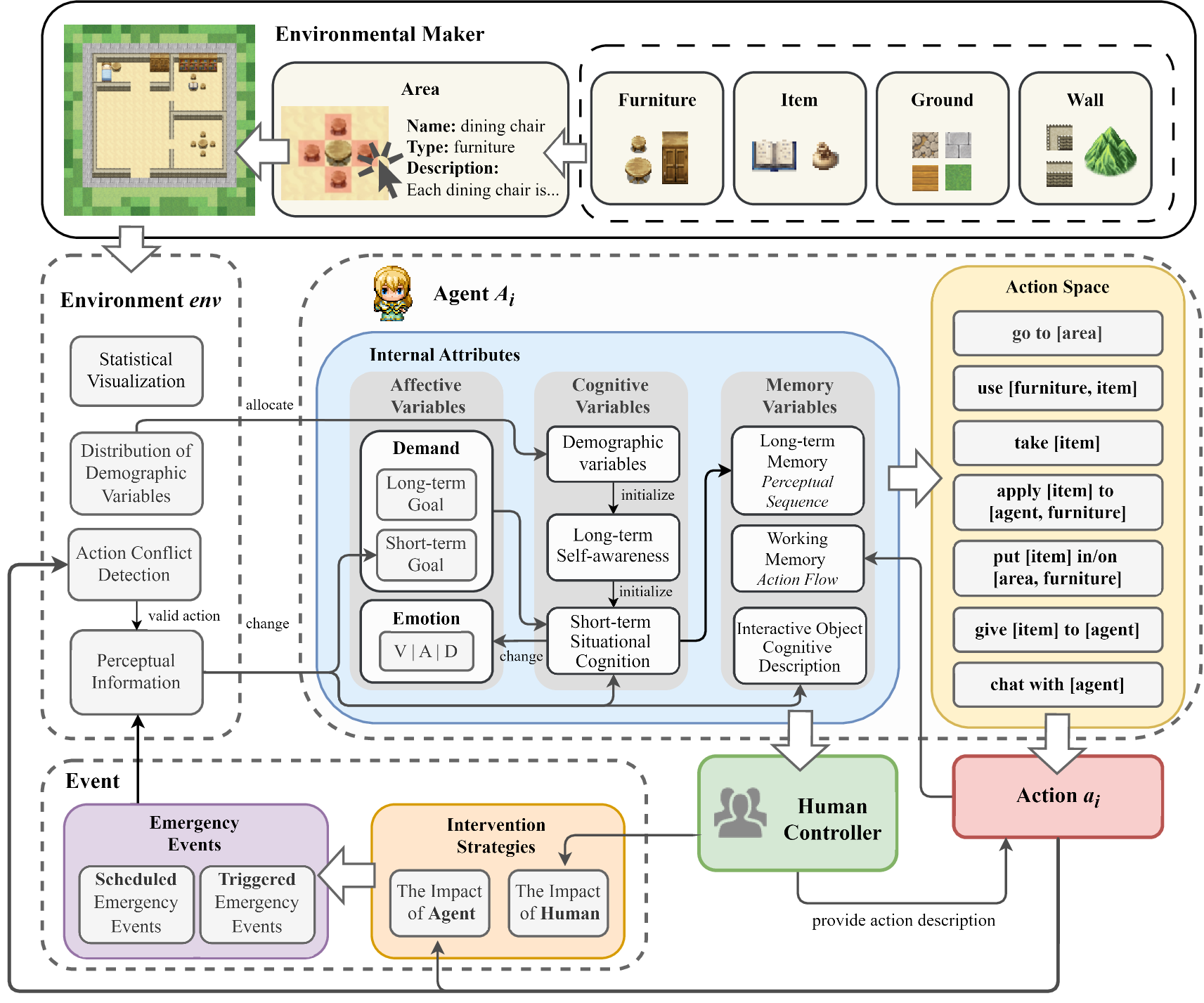}
\caption{Overall Structure of CMASE.}\label{fig1}
\end{figure}

As introduced in Section~\ref{sec1}, CMASE treats the simulation as a simulated ethnographic field in which the researcher acts as an embedded participant; Section~\ref{subsec3.5} formalizes the roles this participant may take. Here, a ``simulated ethnographic field'' refers to an artificial field site in which selected practices of field inquiry, including observation, interviewing, participation, and reflexive adjustment to unfolding interactions, are operationalized within a generative simulation \cite{verbuyst2026artificial}. We use the term to characterize this mode of researcher engagement rather than to imply epistemic equivalence with conventional ethnographic fieldwork. Drawing on natural language descriptions similar to those used in tabletop role-playing games (TRPGs, a game genre in which players use only dialogue and dice to drive imagined adventures) \cite{white2024tabletop}, the framework supports the creation of highly customizable social simulation environments. It facilitates the exploration of how simple interactions among multiple agents give rise to complex collective behaviors within simulated societies. As shown in Figure~\ref{fig1}, the overall structure consists of four components: environment maker, environment, agent, and event. To facilitate practical application and reproducibility, we provide the complete source code, detailed setup instructions, and a step-by-step example for initializing new experiments in our open-source repository: \url{https://github.com/armihia/CMASE}.%\url{https://anonymous.4open.science/r/CMASE-B668}. 

\subsection{Environment Maker}\label{subsec3.1}

We hope that CMASE can support users to define a simulated social environment quickly and simply. To this end, we designed the environment maker module. This module provides a map editor with a basic unit of one grid cell, representing the space occupied by a single agent. According to TRPG conventions, each cell corresponds to approximately five feet in the real world. Creators can place textures anywhere on the map. These textures typically fall into four categories: ground, wall, furniture, and item. Ground textures serve as inert backgrounds. Wall textures function as impassable obstacles and block an agent's perception beyond them. Furniture textures are interactive but fixed in place. Item textures can be picked up, used, or exchanged.

Using textures, creators can define areas on the map. There are two types of areas: region areas and object areas. Each area consists of a collection of textures. A region area is a set of ground textures and includes a description of that region. An object area is a set of furniture or item textures. To support realistic social simulations, the CMASE framework provides an interface for creators to assign functional attributes that align with daily usage scenarios. Specifically, creators can define how an object alters an agent's state based on its real-world utility. For example, reflecting common daily routines, a creator might configure a chair to add a “feeling comfortable” state to an agent, or a repair tool to remove a “damaged” state from a table.

\subsection{Environment}\label{subsec3.2}

The environment module uses the map created by the environment maker as its base. Based on this foundation, researchers can define the distribution of demographic variables for agents to batch-generate generative agents that match the research sample. Researchers may also assign distinct long-term goals (e.g., “maximize social connections within the community,” “accumulate specific resources,” or “maintaining a low profile”) and initial items to agents according to their demographic characteristics. These goals serve as persistent motivational drivers influencing agent behavior throughout the simulation.

Additionally, the module can provide agents with perceptual information and action conflict detection in image and text form, and it can visualize environmental data as statistical charts. At the start of each round, the environment calculates each agent's perceptual information based on their current grid location and line-of-sight obstructions (e.g., walls). It then synthesizes visible environmental data—including nearby terrain, interactive objects, other agents' presence and visible actions, and recent local events—into structured textual descriptions (and optional visual representations). This synthesized data forms the contextual prompt for the agent's cognitive module during decision-making.

Time in the environment advances in rounds. To ensure balanced agent participation and maintain manageable simulation steps, we adopted constraints inspired by traditional tabletop role-playing game (TRPG) systems, including Dungeons \& Dragons (D\&D). These systems provide time-tested heuristics for discretizing continuous human actions into manageable rounds. Consequently, one round corresponds to approximately fifteen seconds in the real world, that is a standard duration in tactical simulations sufficient for completing a single, meaningful interaction. Within a single round, each agent may perform one standard action and one movement action. Movement is limited to a maximum of twenty grid cells per round to mimic realistic movement speeds (approx. 30-60 ft) relative to the map scale. If the standard action is “chat with [agent]” (see Section~\ref{subsubsec3.3.4}), the chat content is limited to thirty words to simulate concise, real-time exchanges and prevent excessive dialogue generation in a single turn.

When agents submit actions to the environment, the system checks for conflicts among actions submitted in the same round, such as multiple agents attempting to pick up the same item. The environment executes the first submitted action and marks conflicting actions as failures, reporting the reason in the next round's perceptual information. 

After all agents have submitted and executed their actions, the environment advances to the next round and provides each agent with updated perceptual information. An exception applies to human interactors, who can select a free action mode for the agents under their control. In free action mode, human interactors are no longer constrained by rounds and can act freely to quantitatively regulate the environment and carry out intervention strategies. In addition, human interactors can take control of any agent in the environment at any time and act from that agent's perspective.

\subsection{Agent}\label{subsec3.3}

The agent module consists of two main components: internal attributes and the action space. Internal attributes are further divided into affective variables, cognitive variables, and memory variables. By receiving perceptual information from the environment, agents can update their internal states and use an LLM to decide on their next actions.

\subsubsection{Affective Variables}\label{subsubsec3.3.1}

To ensure that agents in CMASE can closely simulate individuals with agency in computational social systems, a new proactive architecture incorporating emotion and desire is needed. In this framework, 'needs' are defined as the agent's internal motivational states, distinct from external demands. These needs are concretized into long-term and short-term goals. Long-term goals are defined by the user during agent generation based on the research objectives and remain unchanged afterward. Short-term goals are automatically updated in each round as the agent generates its next action.

In psychological research, Ekman's basic emotion model has been criticized for its coarse categorization and static view of emotions. Recent research draws on the phenomenological concept of “intentionality” to propose a restructured model of emotion. This model represents emotional states as principal component vectors rather than discrete outward expressions, such as happiness, and treats emotion as an internal representation shaped by external stimuli through self-reflection. Accordingly, agents' emotions in CMASE are modeled using the VAD emotion framework \cite{russell1977evidence}. 

Specifically, we utilize the NRC VAD Lexicon v2 \cite{mohammad2025nrc} to map the textual description of the agent's short-term situational cognition to a quantitative emotional vector. The process involves two steps. First, the input text is preprocessed using lemmatization and maximum matching to align words and phrases with the lexicon. The agent's current state vector $\mathbf{E}_t = [v, a, d]$ is then calculated as the arithmetic mean of the Valence, Arousal, and Dominance scores of all identified lexical units in the text. Second, to bridge the gap between continuous numerical vectors and the LLM's linguistic reasoning, we employ a threshold-based discretization method. The continuous range of each dimension is divided into five equidistant intervals, mapping the numerical scores to natural language descriptors: \textit{Low}, \textit{Relatively Low}, \textit{Medium}, \textit{Relatively High}, and \textit{High}. These descriptors are explicitly embedded into the agent's prompt to dynamically constrain and modulate its subsequent reasoning and action generation.

\subsubsection{Cognitive Variables}\label{subsubsec3.3.2}

Cognitive variables include an agent's demographic variables, long-term self-awareness, and short-term situational cognition. Demographic variables are sampled directly from the overall distribution provided by the environment and are used during initialization to generate the agent's long-term self-awareness. These two long-term properties do not change after initialization to preserve the consistency of the agent's personality.

In each round, when the agent makes a decision, the LLM generates new short-term situational cognition based on the agent's long-term self-awareness and the current context. This cognition is analyzed to produce VAD emotional values, which are then converted into a natural language description. This description is fed back to the agent to constrain the direction of its subsequent actions. Essentially, the emotional description dictates the logical boundaries for the agent's behavior. For example, if the VAD values describe a 'furious' state, the agent is constrained to generate impulsive actions rather than rational responses. In addition, when an agent's state changes due to actions or events, its state is appended to the short-term situational cognition so that it can be perceived by the agent itself.

\subsubsection{Memory Variables}\label{subsubsec3.3.3}

Memory variables consist of working memory, long-term memory, and object memory. Working memory is a fixed-length list of recent actions. Whenever an agent executes an action, that action is added to working memory. Likewise, if an agent becomes the target of another agent's action or is affected by an event, the corresponding information is also recorded in its working memory.

Short-term situational cognition and its associated actions are stored in long-term memory in chronological order and are encoded as sentence vectors. When an agent receives new perceptual information, the long-term memory module computes the semantic distance between the incoming short-term cognition and stored memories, then supplies the most relevant linked memories to the LLM in sequence. At the same time, the agent's object memory for any interactable agents, items, or furniture is provided. Object memory entries are generated by the LLM when the agent decides to interact with those objects.

\subsubsection{Action Space}\label{subsubsec3.3.4}

After an agent receives information from perceptual inputs and its internal attributes, it structures that information and passes it to the LLM to determine the next action. There are seven action types available: (1) “\textbf{go to [area]},” a movement action that employs the $A^*$ algorithm to compute an optimal path to a target location within the agent's perceptual range; (2) “\textbf{use [furniture, item]},” which causes the agent to employ the specified object; (3) “\textbf{apply [item] to [agent, furniture]},” by which the agent uses an item to interact with another agent or furniture; (4) “\textbf{take [item]},” enabling the agent to acquire an object; (5) “\textbf{put [item] in/on [area, furniture]},” allowing the agent to place an object into or onto a designated location or piece of furniture; (6) “\textbf{give [item] to [agent]},” through which the agent transfers a held object to another agent; and (7) “\textbf{chat with [agent]},” whereby the agent initiates a dialogue with one or more peers, with the content of the dialogue generated at decision time. 

The action (1) is a movement action, while the remaining actions count as standard actions. Actions (2) and (3) modify the state of both the acting agent and the target, and actions (3), (6), and (7) are recorded in the working memory of the agent being interacted with.

\subsection{Event}\label{subsec3.4}

Emergency events are explicitly designed by researchers to mirror real-world crises and specific social scenarios required for the simulation. When predefined conditions are met, these pre-configured events are activated to change the state of objects within a specified range, defined either by coordinates or by an area. Emergency events are classified as either scheduled or triggered. Scheduled emergency events activate automatically after a set number of rounds. Triggered emergency events activate in two ways: an existence trigger fires when the environment detects an object of a certain type or description, and an action trigger fires when an agent performs a certain action.

Events can fire in sequence: the state changes caused by one event may satisfy the conditions of another event, causing a chain reaction. Through such chains of events, the system can activate special intervention events that modify the short-term goals of selected agents, thereby delivering targeted strategies in response to emergency events. This mechanism makes it possible to observe how agents handle evolving emergency events.

\subsection{Researcher Roles and an Intervention Typology}
\label{subsec3.5}

A recurring critique of human-in-the-loop simulation is that human intervention remains undertheorized: what kinds of intervention exist, through which mechanisms they affect the system, and how they differ from parameter manipulation in conventional ABM. We address this by adapting a classic framework from field research, Gold's spectrum of participant-observation roles ranging from complete observer to complete participant \cite{gold2017roles}, to the computational setting. This yields a four-level typology of researcher engagement, and each level maps onto concrete CMASE mechanisms.

\textbf{Level 1, Observer.} The researcher monitors the simulation without entering it. Engagement is limited to the environment module's statistical visualizations, trajectory data, and logs. No agent perceives the researcher, and the system remains causally closed. This level corresponds to the use mode of existing GABM platforms.

\textbf{Level 2, Interviewer.} The researcher addresses agents through structured or semi-structured dialogue. Interview-style querying is available in several prior platforms \cite{piao2025agentsociety, tang2024gensim, zhang2025socioverse}; in those systems, however, the interview operates as a side channel outside the simulated world, designed so that agents respond without their ongoing behavior being affected. In CMASE, an interview is a social event within the world: it enters the interviewed agent's working memory and may shift its short-term situational cognition and VAD emotional state through the same perceptual pipeline as agent-to-agent interaction (Section \ref{subsec3.3}), while the physical environment and other agents remain unaffected. This level is the computational analogue of ethnographic interviewing.

\textbf{Level 3, Participant.} The researcher takes control of an agent and acts from its perspective under the same round-based constraints as all other agents: one standard action and one movement action per round, the thirty-word chat limit, and conflict detection (Section \ref{subsec3.2}). The researcher is thus a peer social actor whose influence propagates only through ordinary social channels such as conversation, object exchange, and co-presence, and whose actions are perceived, remembered, and emotionally appraised by other agents exactly as any agent's actions would be.

\textbf{Level 4, Intervener.} The researcher enters free action mode, suspending round constraints to enact environment-level changes: triggering events, modifying object states, or assigning intervention strategies that alter selected agents' short-term goals (Section \ref{subsec3.4}). This level models exogenous shocks and policy instruments. Even here, agents experience the changes through their perceptual pipeline rather than through silent rewrites of their internal variables, so downstream reactions remain endogenous.

This typology clarifies the distinction between CMASE and parameter manipulation in conventional ABM. Parameter manipulation is exogenous, static, and specified in advance: it alters an agent's state from outside, with no event that any other agent perceives as a social act. CMASE interventions at Levels 2 to 4 are endogenous social events: they occur within the run, are mediated by agents' perception, memory, and affect, and can therefore be resisted, reinterpreted, or amplified by agents—where such amplification may carry the intervention forward or, equally, consolidate resistance to it. The intervention is not a boundary condition of the model but a modeled social variable, which is the defining requirement of comprehensive modeling identified at the opening of this section. This exogenous parameter manipulation is therefore not a level of the typology but the limiting case against which the typology defines itself; in Section~\ref{sec5} we instantiate it as the parametric condition and contrast it directly with a Level 3 embedded intervention.

An additional distinction is important here. The difference between embedded participation and direct parameter manipulation concerns how an intervention enters the simulation, rather than whether the intervening actor is human. A preconfigured or automated agent could also produce an identifiable in-world social event. Human control adds a further capability. The researcher can observe the unfolding interaction, interpret agents' responses, and adjust subsequent actions during the same run, while those actions continue to enter the simulation through ordinary in-world social channels and remain attributable to an identifiable source. This adaptive participation is an affordance of CMASE rather than what is isolated in the controlled comparison of Section 5.4. There, the intervention protocol is intentionally standardized so that adaptive participation is not itself the varying factor, allowing the comparison to isolate embedded versus parametric delivery, including their differences in social attribution and process traceability.

\subsection{Reusability and Implementation Cost}
\label{subsec3.6}

A practical question for any simulation platform is how much must be rewritten to study a new phenomenon. In CMASE the four components of this section are fixed infrastructure, and a study is specified largely through configuration: an environment map built with the environment maker, a population generated by the environment module from a demographic distribution and a set of long-term goals, and, where an intervention is studied, a choice of researcher role from the typology in Section~\ref{subsec3.5}. The cognitive, affective, and memory machinery of the agent, the round structure, the action space, and the event system are not touched.

This design has a consequence that matters for the experiments in Section~\ref{sec5}. The conditions compared there, the no-intervention baseline, a demonstration-only control, a parametric intervention, and an embedded intervention, run on one shared specification: the same village cast and demographic variables, the same three contribution dispositions, the same payoff accounting, and the same logging. The conditions differ only in how an intervention reaches the agents. The parametric condition adds the intervener's message to the targeted agents' short-term situational cognition through a single injected field, with no in-world event, which is the limiting case described at the end of Section~\ref{subsec3.5}. The embedded condition adds nothing to that field; instead it uses the Level~3 participant role, so the same message and the same visible contribution arrive as the ordinary movement and speech of a present agent and pass through the perceptual pipeline of Section~\ref{subsec3.3}. Moving from one to the other is a change at the point of delivery rather than a reimplementation of the scenario.

Two properties follow. First, the engineering cost of a new condition or a new study is low, because it reuses the shared infrastructure rather than rebuilding it. Second, and more important for the claims of this paper, holding the surrounding specification fixed is what makes the comparison meaningful: because the cast, the dispositions, the payoffs, and the logging are identical across conditions, an observed difference between parametric and embedded delivery can be interpreted as an effect of the delivery channel within this controlled simulation comparison, rather than as a consequence of incidental differences in how each condition was built. The same property that lowers implementation cost is therefore also what licenses the controlled contrast in Section~\ref{sec5}. The repository linked above provides the full configuration for each study so that these specifications can be inspected and rerun.

\section{Platform Demonstration: Simulating Context-Dependent Social Dynamics}\label{sec4}

Our empirical evaluation examines complementary aspects of CMASE across two studies. Table~\ref{tab:evaluation_roadmap} summarizes the capability examined in each evaluation, its comparison or research question, and the principal measures and outcomes.

\begin{table*}[t]
\centering
\caption{\textbf{Evaluation roadmap of CMASE.}
Each evaluation is mapped to the CMASE capability examined, the comparison or research question, and its principal outcome measures and key findings.}
\label{tab:evaluation_roadmap}

\small
\setlength{\tabcolsep}{3.5pt}
\renewcommand{\arraystretch}{1.12}

\begin{tabularx}{\textwidth}{
@{}
>{\centering\arraybackslash}p{0.040\textwidth}
>{\raggedright\arraybackslash}p{0.115\textwidth}
>{\raggedright\arraybackslash}p{0.145\textwidth}
>{\raggedright\arraybackslash}p{0.200\textwidth}
>{\raggedright\arraybackslash}X
@{}
}
\toprule
\textbf{Sec.}
&
\textbf{Evaluation}
&
\textbf{CMASE capability}
&
\textbf{Comparison or RQ}
&
\textbf{Principal measures and key outcomes}
\\
\midrule

% ============================================================
% Study 1
% ============================================================

\multicolumn{5}{@{}l}{%
\textbf{\textit{Study 1: Green Space and Social Fragmentation}}}
\\
\midrule

% ------------------------------------------------------------
% 4.3
% ------------------------------------------------------------

\ref{subsec4.3}
&
Quantitative reconstruction
&
Pattern reproduction
&
Does CMASE reproduce the reported vegetation effect?
&
\begin{minipage}[t]{\linewidth}
\vspace{0pt}
\footnotesize
\setlength{\tabcolsep}{2pt}

\begin{tabular}{
@{}
>{\raggedright\arraybackslash}p{0.50\linewidth}
>{\centering\arraybackslash}p{0.20\linewidth}
>{\centering\arraybackslash}p{0.20\linewidth}
@{}
}
\textbf{Indicator}
& \textbf{Low}
& \textbf{High}
\\
\cmidrule(lr){2-3}

Distrust$^{*}$
& 4.60
& 3.60
\\

Exploitation$^{*}$
& 4.30
& 3.40
\\

Indifference
& 3.70
& 2.70
\\
\end{tabular}
\end{minipage}
\\

\addlinespace[3pt]

% ------------------------------------------------------------
% 4.4
% ------------------------------------------------------------

\ref{subsec4.4}
&
Embedded inquiry
&
In-world inquiry
&
Can embedded inquiry reveal process-level patterns behind the aggregate result?
&
\begin{minipage}[t]{\linewidth}
\vspace{0pt}
\footnotesize
\setlength{\tabcolsep}{2pt}

\begin{tabular}{
@{}
>{\raggedright\arraybackslash}p{0.48\linewidth}
>{\raggedright\arraybackslash}p{0.44\linewidth}
@{}
}
\textbf{Indicator}
& \textbf{Observed pattern}
\\

Indoor-agent coping
& Occupation-dependent coping
\\

Green-space use
& Lakeside clustering
\\
\end{tabular}
\end{minipage}
\\

\addlinespace[3pt]

% ------------------------------------------------------------
% 4.5
% ------------------------------------------------------------

\ref{subsec4.5}
&
Agent and record evaluation
&
Fidelity and interpretability
&
Are assigned personas preserved and do the records support interpretation?
&
\begin{minipage}[t]{\linewidth}
\vspace{0pt}
\footnotesize
\setlength{\tabcolsep}{2pt}

\begin{tabular}{
@{}
>{\raggedright\arraybackslash}p{0.50\linewidth}
>{\centering\arraybackslash}p{0.40\linewidth}
@{}
}
\textbf{Indicator}
& \textbf{Mean score}
\\

Persona fidelity
& 4.23
\\

Interpretive affordances
& 4.25
\\
\end{tabular}
\end{minipage}
\\

\addlinespace[3pt]

% ------------------------------------------------------------
% 4.6
% ------------------------------------------------------------

\ref{subsec4.6}
&
Runtime usability
&
Real-time participation
&
How does runtime scale, and what is the real-time participation bound?
&
\begin{minipage}[t]{\linewidth}
\vspace{0pt}
\footnotesize
\setlength{\tabcolsep}{2pt}

\begin{tabular}{
@{}
>{\raggedright\arraybackslash}p{0.56\linewidth}
>{\centering\arraybackslash}p{0.34\linewidth}
@{}
}
\textbf{Indicator}
& \textbf{Outcome}
\\

Step time (1--10 agents)
& 23 s
\\

Real-time participation bound
& $\leq 100$ agents
\\
\end{tabular}
\end{minipage}
\\

% ============================================================
% Study 2
% ============================================================

\midrule
\multicolumn{5}{@{}l}{%
\textbf{\textit{Study 2: Public-Goods Cooperation and Intervention}}}
\\
\midrule

% ------------------------------------------------------------
% 5.2
% ------------------------------------------------------------

\ref{subsec5.2}
&
Sealed condition
&
Dynamic reproduction
&
Does CMASE reproduce the published cooperation dynamics?
&
\begin{minipage}[t]{\linewidth}
\vspace{0pt}
\footnotesize
\setlength{\tabcolsep}{2pt}

\begin{tabular}{
@{}
>{\raggedright\arraybackslash}p{0.56\linewidth}
>{\centering\arraybackslash}p{0.34\linewidth}
@{}
}
\textbf{Indicator}
& \textbf{Outcome}
\\

Groups crossing both trajectory bands
& 2 of 26
\\

Early--endpoint correlation
& $\rho=0.90$
\\
\end{tabular}

\vspace{2pt}

\begin{tabular}{
@{}
>{\raggedright\arraybackslash}p{0.50\linewidth}
>{\centering\arraybackslash}p{0.20\linewidth}
>{\centering\arraybackslash}p{0.20\linewidth}
@{}
}
\textbf{Indicator}
& \textbf{Low state}
& \textbf{High state}
\\
\cmidrule(lr){2-3}

State persistence
& 1.00
& 0.95
\\
\end{tabular}

\end{minipage}
\\

\addlinespace[3pt]

% ------------------------------------------------------------
% 5.3
% ------------------------------------------------------------

\ref{subsec5.3}
&
Open condition
&
Context-sensitive dynamics
&
Does social interaction alter cooperation dynamics?
&
\begin{minipage}[t]{\linewidth}
\vspace{0pt}
\footnotesize
\setlength{\tabcolsep}{2pt}

\begin{tabular}{
@{}
>{\raggedright\arraybackslash}p{0.50\linewidth}
>{\centering\arraybackslash}p{0.20\linewidth}
>{\centering\arraybackslash}p{0.20\linewidth}
@{}
}
\textbf{Indicator}
& \textbf{Sealed}
& \textbf{Open}
\\
\cmidrule(lr){2-3}

Endpoint cooperation (0H, 2C, 3L)
& 0.87
& 0.00
\\

Endpoint cooperation (1H, 2C, 2L)
& 2.03
& 0.73
\\

Contribution-directed action share
& 0.18
& 0.07
\\

Home-directed action share
& 0.21
& 0.41
\\
\end{tabular}

\end{minipage}
\\

\addlinespace[3pt]

% ------------------------------------------------------------
% 5.4
% ------------------------------------------------------------

\ref{subsec5.4}
&
Intervention delivery
&
Embedded intervention and traceability
&
How do parametric and embedded delivery differ?
&
\begin{minipage}[t]{\linewidth}
\vspace{0pt}
\footnotesize
\setlength{\tabcolsep}{2pt}

\begin{tabular}{
@{}
>{\raggedright\arraybackslash}p{0.50\linewidth}
>{\centering\arraybackslash}p{0.20\linewidth}
>{\centering\arraybackslash}p{0.20\linewidth}
@{}
}
\textbf{Indicator}
& \textbf{P}
& \textbf{E}
\\
\cmidrule(lr){2-3}

Intervention-window cooperation
& 2.33
& 2.33
\\

In-world social record
& No
& Yes
\\

Source verifiability
& No
& Yes
\\
\end{tabular}

\end{minipage}
\\
\\

\bottomrule
\end{tabularx}

\vspace{2pt}

\begin{minipage}{\textwidth}
\footnotesize
$^{*}$ indicates a statistically significant between-condition comparison
with $p<.05$ in Section~\ref{subsec4.3}. Scores in Section~\ref{subsec4.5} are reported on a 1--5 scale.
P = parametric intervention; E = embedded intervention.
\end{minipage}

\end{table*}

To evaluate CMASE in a published social-scientific setting, we conduct a computational replication of a field study examining the relationship between urban greenery and perceived social fragmentation \cite{lee2025greenery}. Here, computational replication refers to reconstructing the source study's setting, variables, and measurement procedure in a generative environment and examining whether the reported pattern re-emerges through situated agent interaction. It is distinct from an independent replication with new human participants or a prospective prediction test. The general positive relationship between green space and social cohesion is well documented and may be represented in the model's pre-training data \cite{wan2021underlying}. We therefore do not interpret directional agreement alone as establishing empirical validity for the corresponding human population. Instead, the evaluation asks whether the reported relationship is reproduced through the reconstructed environment while CMASE additionally provides trajectories, affective dynamics, interviews, and embedded observation through which the generated process can be examined.

Although Lee and Han appeared in an urban-sustainability venue, the research question and its constructs are social-scientific rather than environmental. The dependent variables are core constructs of sociology and social psychology, and the underlying data are qualitative interviews analyzed for social meaning. The relationship between urban green space and social cohesion is itself an established topic in environmental sociology and social-capital research \citep{wan2021underlying}. What CMASE computationally replicates here is the reported relationship between a shared environment and perceived community fragmentation, not a biophysical process. The public goods study in Section~\ref{sec5} complements this case with a setting drawn from experimental economics, whose social-scientific character is not in question.

\subsection{Setup}\label{subsec4.1}

This is a retrospective computational replication: the source study and its reported outcomes were known when the simulation was designed and are therefore comparison targets rather than held-out predictions. We used GPT-4o ``2024-12-01-preview'' for a study published after that model version to reduce the narrower possibility of direct memorization of the source paper. This does not exclude broader prior knowledge about green space and social cohesion encoded in the model's training data. The source study informs the vegetation manipulation, the isolation context, and the three outcome constructs and corresponding questionnaire. Agent personas and long-term goals are held fixed across the high- and low-vegetation conditions; the treatment difference is introduced through the environment rather than condition-specific persona instructions. The generated trajectories, affective states, interactions, and questionnaire responses are then compared across these conditions.

Lee and Han \cite{lee2025greenery} analyzed interview data collected in South Korea during the COVID-19 pandemic and demonstrated that regions with higher vegetation cover exhibited lower levels of social fragmentation, indicating stronger community bonds. Based on the descriptions in the original study, we used the environment maker module to construct maps of South Korean regions with varying vegetation cover. Because the original paper emphasized South Korea's community isolation policies during the pandemic, we designed an isolated community environment and placed 10 agents in it. The population of ten mirrors the micro-scale of the replicated setting and lies well within the usability bound for real-time human embedding (Section~\ref{subsec4.6}), because embedding a researcher into the simulation requires acting within the round structure. We treat population size as set by the social setting being modeled rather than as a parameter to be varied, since both this community and the small-group setting of Section~\ref{sec5} are micro-scale by construction.

Specifically, the simulated community comprises ten individual residential units surrounding a shared public green space. To support diverse activities, the green space was designed with distinct functional zones, including a recreation area, a rest zone, and a scenic lakeside area equipped with benches. We placed ten agents in this environment, each initialized with a distinct persona, including a specific name, age, and occupation (e.g., lawyer, fashion designer, software engineer), to simulate a heterogeneous micro-society.

The four components of Section~\ref{sec3} map onto this scenario as follows: the environment maker builds the community map and operationalizes vegetation level as the independent variable; the environment module generates ten agents with occupation-consistent goals; the pandemic isolation policy is realized as a standing spatial constraint that confines agents to the community; and the researcher-embedding interface (Section~\ref{subsec3.5}) is used at the Participant and Interviewer levels, where one author entered the field as an agent and conducted Level 2 in-world interviews. Implementation details for each component, full parameter settings, and the complete list of agent long-term goals are given in Appendix~\ref{app:config}.

\subsection{Measures}\label{subsec4.2}

In the original study, higher levels of vegetation were associated with lower social fragmentation, operationalized through distrust, exploitation, and indifference. Using CMASE’s environment builder, we recreated regional conditions with varying vegetation levels and placed 10 agents in each simulated environment. 

After running the simulation for 50 time steps to reach equilibrium, we interviewed each agent using a structured Likert-scale questionnaire adapted from the original study \cite{lee2025greenery} to measure their sense of social fragmentation, which comprises three dimensions: distrust, exploitation, and indifference. In addition, agents completed a 7-point Likert scale questionnaire corresponding to these three subdimensions. The subdimensions are defined as follows: (a) Distrust denotes skepticism or lack of confidence in the integrity and functionality of social structures. (b) Exploitation refers to the perception or experience of being used unjustly or immorally for another's benefit in a fragmented society. (c) Indifference manifests as a lack of interest, concern, or empathy for social issues or the plight of others. 

As additional process measures, we analyzed the agents' movement trajectories and emotional dynamics across different regions. We defined each agent's overall emotional value as the sum of its VAD emotion vector components and calculated the ambient mood of each region as the mean emotional value of all agents within that region.

\subsection{Quantitative Replication}\label{subsec4.3}

In the original study, higher levels of vegetation were associated with lower social fragmentation. We observed that the CMASE simulation successfully reproduces this decline (Table \ref{tab:cmase_human_comparison}). As vegetation cover increases, the mean scores for distrust fall from \(4.60 \pm 0.97\) to \(3.60 \pm 1.17\), for exploitation from \(4.30 \pm 1.34\) to \(3.40 \pm 1.43\), and for indifference from \(3.70 \pm 1.16\) to \(2.70 \pm 1.49\). This indicates that higher vegetation indices correspond to a statistically significant decrease in agents' sense of social fragmentation.

\begin{table}[t]
\centering
\caption{Comparison of CMASE simulation results with human baseline.}
\begin{tabular}{lcccccc}
\toprule
 & \multicolumn{2}{c}{CMASE} & \multicolumn{2}{c}{Human NDVI} & \multicolumn{2}{c}{Human EVI} \\
\cmidrule(lr){2-3} \cmidrule(lr){4-5} \cmidrule(lr){6-7}
Dimension & Low Veg & High Veg & Low Veg & High Veg & Low Veg & High Veg \\
\midrule
Distrust     & $4.60 \pm 0.97$ & $3.60 \pm 1.17$ & 4.0 & 1.6 & 4.5 & 1.1 \\
Exploitation & $4.30 \pm 1.34$ & $3.40 \pm 1.43$ & 5.0 & 1.4 & 4.5 & 1.6 \\
Indifference & $3.70 \pm 1.16$ & $2.70 \pm 1.49$ & 5.1 & 1.3 & 5.5 & 0.9 \\
\bottomrule
\end{tabular}
\label{tab:cmase_human_comparison}
\end{table}

Paired-sample t-tests confirmed significant decreases in distrust (\(t = 2.739, p = 0.023\)) and exploitation (\(t = 3.250, p = 0.010\)), while the decrease in indifference was not statistically significant (\(t = 1.793, p = 0.107\)). 

Effect size analyses with 95\% confidence intervals further supported these findings, with reductions in distrust (d=0.93, 95\% CI [0.20, 1.60]) and exploitation (d=0.65, 95\% CI [0.40, 1.40]) showing moderate to large effects. In contrast, the effect on indifference was less certain (d=0.75, 95\% CI [-0.10, 2.00]).

\begin{figure}[h]
\centering
\includegraphics[width=0.9\textwidth]{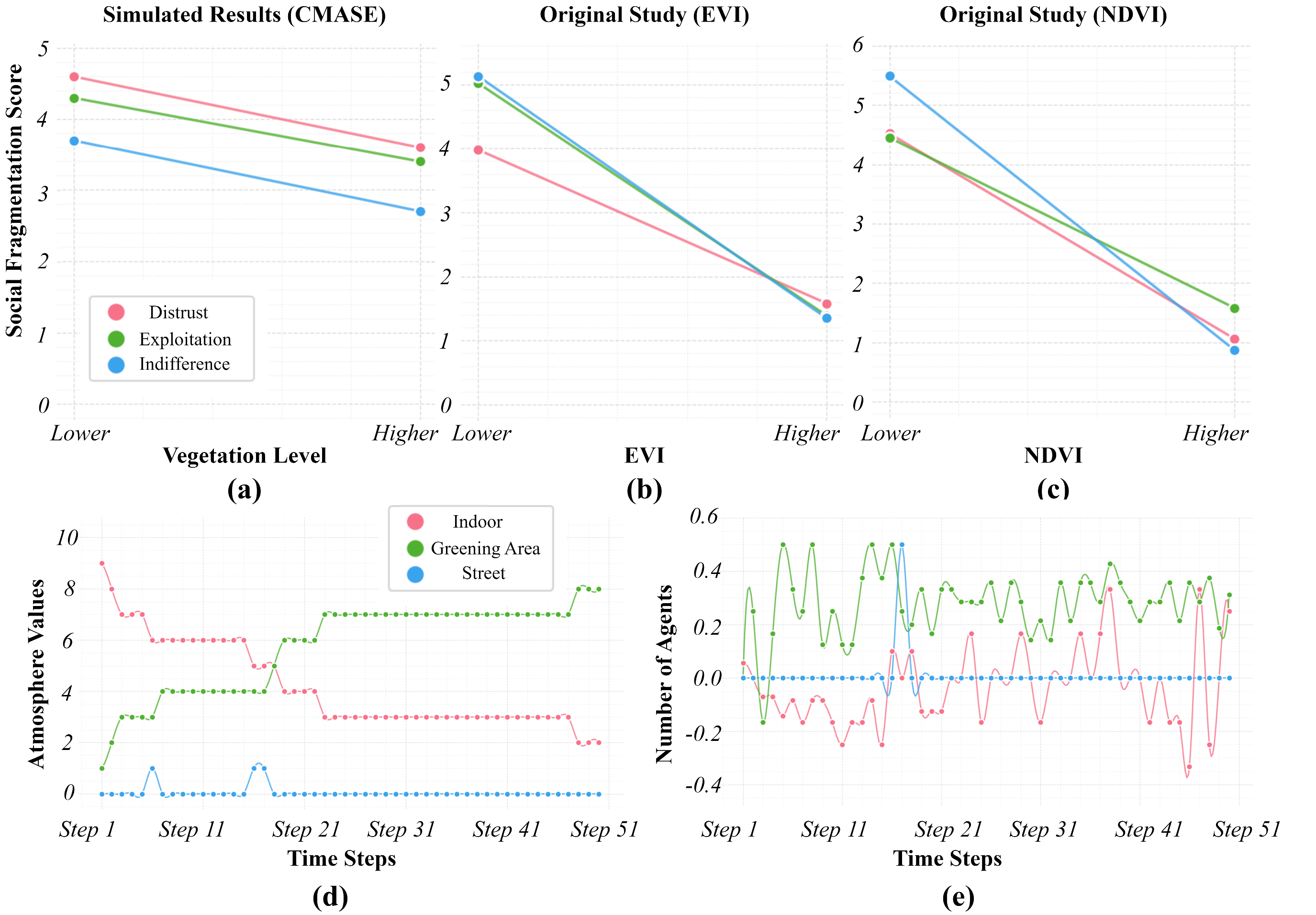}
\caption{Comparison of CMASE simulation with the original study results. (a) Simulation results generated by the CMASE framework. (b) Predictions using the EVI index from the original study. (c) Predictions using the NDVI index from the original study. (d) Movement trajectories of agents during CMASE simulation. (e) Regional variation in agents’ emotional states.}\label{fig-exp1}
\end{figure}

For comparison, the original study used two indicators, the Enhanced Vegetation Index (EVI) and the Normalized Difference Vegetation Index (NDVI), to quantify vegetation levels. As EVI and NDVI increased, scores of distrust (from 4.0 and 4.5 to 1.6 and 1.1), exploitation (from 5.0 and 4.5 to 1.4 and 1.6), and indifference (from 5.1 and 5.5 to 1.3 and 0.9) all exhibited a decreasing trend (Figure \ref{fig-exp1} a-c).

The results from CMASE clearly show the suppressive effect of vegetation levels on the sense of social fragmentation. This is a key phenomenon identified in the study by Lee and Han. It is important to note that the CMASE simulation excluded factors that were not considered or difficult to exclude in the original study, such as regional culture, which are also related to the sense of social fragmentation. Therefore, compared to the larger decreases reported in the original study, the changes observed in our simulation are relatively modest. This indicates that our simulation results are largely consistent with the original findings.

We also analyzed agent trajectories and affective states. Movement data showed that agents increasingly leave indoor areas and move toward vegetated regions, with the ambient mood in vegetated regions consistently remaining higher than that in indoor areas. This pattern aligns with the original study's discussion and demonstrates CMASE's ability to simulate collective social behavior in social science research.

\subsection{Artificial Fieldwork}\label{subsec4.4}

In addition to the quantitative replication, we entered the CMASE environment as a co-resident participant and interacted with the other agents during the simulation. In doing so, we exercise two levels of the intervention typology of Section~\ref{subsec3.5}: the Level 2 interviewer role, in querying agents directly, and the Level 3 participant role, in entering the field as a co-resident agent. Through this embedded inquiry, we examine process-level patterns that complement the aggregate findings reproduced from the original study. We conducted interview analyses with two agents who remained indoors throughout the experiment (Figure \ref{fig-exp1} d-e). Full interview transcripts and detailed analyses are provided in Appendix~\ref{app:ethnography}. The results indicate that agents with different occupations exhibited varying mechanisms for mitigating the sense of social fragmentation in isolation. Agents in creative and analytical professions tended to rely more on orderly and low-stimulus environments, using these conditions to construct an internally stable and distinct mental world that anchored their cognitive states. For these agents, the green space had a relatively limited effect in alleviating their sense of social fragmentation.

In contrast, agents who actively moved to the green space tended to gather around the lakeside bench area rather than in the predefined activity zones. According to interview data, this non-programmed choice suggests that they were seeking not only social interaction but also a form of self-experience. This pattern is similar to embodied experiences in humans, and such ‘experiential-like’ mechanisms seem to play a role in linking green space with feelings of social fragmentation. This behavior may reflect an “embodied-like” tendency within the simulation. We remain cautious in interpreting this trend, as it could also stem from hallucinatory processes.

From a language-philosophical perspective, linguistic structures and experiential structures may exhibit a certain degree of partial isomorphism or structural correspondence \cite{sun2024can, Haspelmath2008, pederson1998semantic}. Thus, LLMs trained on human language may partially acquire patterns of human experience (although this does not imply that they possess or actually experience these phenomena). While we remain cautious in interpreting this “embodied-like” tendency, as it could also stem from hallucinatory processes, it is equally reasonable that agents, by design, prioritize coherence with their predefined human personas. For a human persona, embodied reflections are integral to an authentic experience; thus, simulating these expressions is a necessary feature of a robust social simulation. In any case, these patterns suggest that agents can generate persona-consistent expressions related to embodied experience within the simulation, while not implying that the agents themselves possess embodied or experiential states.

\subsection{Evaluation on Agent Persona and Ethnographic Affordances}\label{subsec4.5}

A natural question for any generative-agent study is whether the agents actually behave as the experimental design specifies, rather than only being configured correctly at initialization. To examine this, we conducted a two-stage evaluation over the interaction logs and interview transcripts from the community isolation experiment, combining an automated LLM-based assessment with independent human coding. The evaluation considered two aspects: whether agents express their assigned personas faithfully throughout the run (Agent Persona Fidelity, Table~\ref{tab:persona_fidelity}), and whether the resulting records support interpretive analysis (Ethnographic Affordances, Table~\ref{tab:ethnographic_affordances}). To reflect the demographic variables integrated into the agent personas within the simulated field described in Section~\ref{subsec4.1}, we categorized the agents into four distinct clusters: Creative and Design (e.g., Fashion Designer), Analytical and Academic (e.g., Lawyer), Technical and Strategic (e.g., Software Engineer), and Public and Healthcare (e.g., Cardiologist).

Our assessment of agent persona quality focused on three dimensions: consistency (alignment with long-term self-awareness), responsiveness (logical feedback to environmental stimuli), and emotional alignment (accuracy of emotional state transitions). For ethnographic affordances, we measured thick description (richness of the linguistic world), traceability (clarity of the decision chain), interaction depth (sense of meaningful dialogue), and mechanistic insight (ability to reveal social logic). We first used Gemini-3.1 Pro to rate each of the ten agent records across these seven dimensions on a five-point scale. To provide an independent human check, two human coders subsequently evaluated the same records using the same dimension definitions and rating scale. The coders were blinded to the Gemini scores and rated the records independently without consulting each other's scores. Human--human agreement was assessed using quadratic-weighted Cohen's \(\kappa\). As a further independent behavioral check, the public goods study in Section~\ref{sec5} reports the rate at which agents act in accordance with their assigned contribution dispositions.

\begin{table}[htbp]
\centering
\caption{LLM-based automated evaluation of Agent Persona Fidelity}
\label{tab:persona_fidelity}
\begin{tabular}{lcccc}
\toprule
\textbf{Persona Cluster} & \textbf{Consistency} & \textbf{Responsiveness} & \textbf{Emotional Alignment} & \textbf{Overall} \\ \midrule
Creative and Design & 4.00 & 3.33 & 4.67 & 4.00 \\
Analytical and Academic & 4.33 & 3.33 & 4.33 & 4.00 \\
Technical and Strategic & 4.00 & 4.00 & 5.00 & 4.33 \\
Public and Healthcare & 5.00 & 4.50 & 5.00 & 4.83 \\ \midrule
\textbf{Average} & \textbf{4.30} & \textbf{3.70} & \textbf{4.70} & \textbf{4.23} \\ \bottomrule
\end{tabular}
\end{table}

\begin{table}[htbp]
\centering
\caption{LLM-based automated evaluation of Ethnographic Affordances}
\label{tab:ethnographic_affordances}
\begin{tabular}{lccccc}
\toprule
\textbf{Persona Cluster} & \textbf{\begin{tabular}[c]{@{}c@{}}Thick\\ Description\end{tabular}} & \textbf{Traceability} & \textbf{\begin{tabular}[c]{@{}c@{}}Interaction\\ Depth\end{tabular}} & \textbf{\begin{tabular}[c]{@{}c@{}}Mechanistic\\ Insight\end{tabular}} & \textbf{Overall} \\ \midrule
Creative and Design & 4.33 & 4.00 & 3.33 & 3.67 & 3.83 \\
Analytical and Academic & 5.00 & 4.00 & 4.33 & 4.00 & 4.33 \\
Technical and Strategic & 4.00 & 4.00 & 4.50 & 4.50 & 4.25 \\
Public and Healthcare & 5.00 & 5.00 & 4.50 & 4.50 & 4.75 \\ \midrule
\textbf{Average} & \textbf{4.60} & \textbf{4.20} & \textbf{4.10} & \textbf{4.10} & \textbf{4.25} \\ \bottomrule
\end{tabular}
\end{table}

To independently assess the automated evaluation, two human coders rated all ten agent records. The coders showed substantial agreement overall (quadratic-weighted Cohen's \(\kappa=0.718\)), with stronger agreement for Ethnographic Affordances (\(\kappa=0.793\)) and moderate agreement for Agent Persona Fidelity (\(\kappa=0.599\)). Across all seven dimensions, the mean human rating was 4.02, compared with 4.24 from Gemini. For the three interpretive dimensions most directly related to ethnographic affordances, the aggregate ratings were closely aligned: thick description received 4.50 from the human coders and 4.60 from Gemini, interaction depth 4.15 and 4.10, and mechanistic insight 4.00 and 4.10, respectively. We therefore treat the Gemini ratings as an automated diagnostic complemented by independent human evaluation rather than as independent validation on their own.

For Agent Persona Fidelity, the automated evaluation yielded an overall score of 4.23, with scores of 4.30 for consistency, 3.70 for responsiveness, and 4.70 for emotional alignment. The more moderate inter-coder agreement observed for this evaluation block suggests that persona-related judgments, particularly those concerning internal emotional alignment, are more evaluator-dependent than the ethnographic-affordance ratings. We therefore interpret these scores primarily as diagnostic evidence that the generated records exhibit persona-related behavioral structure, rather than as a definitive validation of individual persona fidelity.

Regarding Ethnographic Affordances, the automated evaluation yielded consistently high scores for thick description (4.60), traceability (4.20), interaction depth (4.10), and mechanistic insight (4.10). The independent human evaluation showed stronger inter-coder agreement for this block (\(\kappa=0.793\)), and the aggregate human ratings were closely aligned with the automated ratings on the interpretive dimensions of thick description, interaction depth, and mechanistic insight. Together, these results suggest that the simulation records provide useful material for ethnography-inspired interpretation, particularly through descriptive richness, process traceability, and sustained interaction. Overall, CMASE provides an interpretable and interactive foundation for embedded inquiry in computational sociology.

\subsection{Time and Cost Analysis}\label{app:time_cost}\label{subsec4.6}

\begin{figure}[h]
\centering
\includegraphics[width=0.9\textwidth]{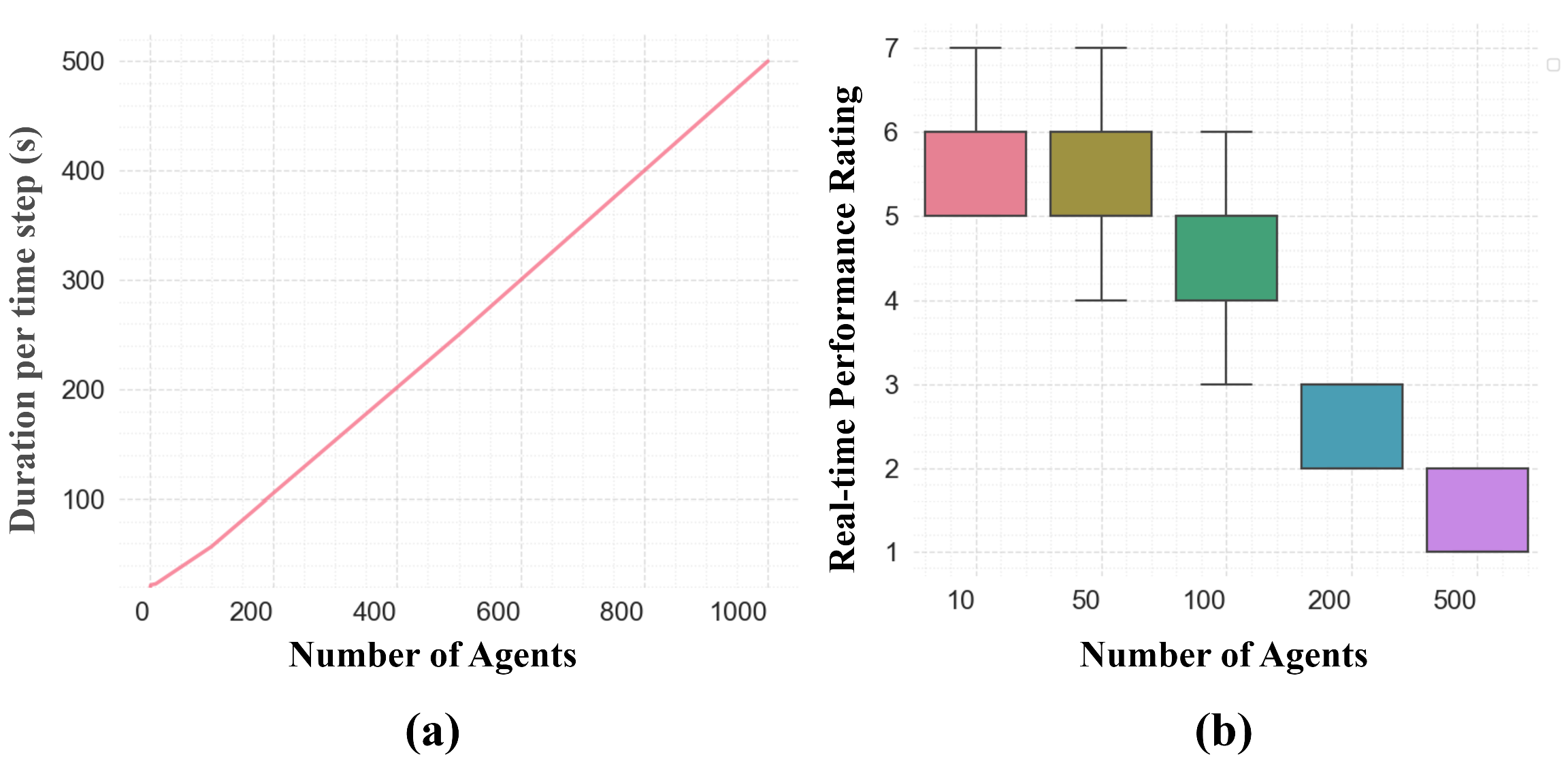}
\caption{CMASE performance under different agent counts. (a) Duration per time step. (b) Human evaluator responsiveness ratings.}\label{fig10}
\end{figure}

To assess the practicality of CMASE, we conducted a human-in-the-loop responsiveness evaluation. Five evaluators were recruited for this study. The selection criteria required participants to be familiar with tabletop role-playing games to ensure they could accurately judge the system's fluidity against genre standards. The evaluators engaged in real-time interactions within environments containing varying numbers of agents. They rated the system's responsiveness on a 7-point Likert scale. The results are shown in Figure \ref{fig10}.

Figure \ref{fig10}(a) illustrates that the time required for a single time step in CMASE grows linearly with the number of agents. With 1 to 10 agents, the average duration is 23s, which includes both the mean API call latency and the environment's action execution time. It increases to 498s when 1,000 agents are present. 

Figure \ref{fig10}(b) presents boxplots of human evaluator scores. The purpose of this evaluation was not merely to assess speed, but to identify the critical threshold for real-time human-in-the-loop usability. Our findings reveal that human evaluators require an average of approximately 50s per time step to process information and select actions. Consequently, as long as the system's calculation time remains below this 50s window (which corresponds to agent counts $\le$ 100), the system latency is effectively masked by the human's own cognitive load, resulting in stable, high responsiveness ratings. However, once the agent count exceeds 100, the system latency surpasses the human operation window, causing waiting time to accrue and responsiveness ratings to drop partially. At 500 agents, the responsiveness ratings hit the measurement floor. Further testing was deemed unnecessary given the diminishing marginal utility of data at already unacceptable latencies, indicating the practical upper bound for real-time human participation in CMASE.

In terms of cost, each agent's action in one time step incurs approximately \$0.10 in API expenses. To put this into perspective, we compared our cost structure with the benchmark set by Park et al. \cite{park2023generative}, who reported spending “several thousand dollars” to simulate 25 agents for two days (about 192–576 time steps). Applying CMASE's rate to a simulation of the same scale would result in an estimated total cost of \$480 to \$1,440. This indicates that CMASE offers a more economically feasible solution for multi-agent simulations of this magnitude.

\section{Computational Replication of Cooperation Dynamics and Testing the Intervention Typology}
\label{sec5}

The green-space study in Section~\ref{sec4} shows that CMASE can reproduce a published environmental-behavioral finding. We now test the framework on a second phenomenon from a different field, the dynamics of cooperation in a repeated public goods game, and we use this setting to build the baselines that the intervention experiments later in this section require. The unit of interest is not an individual's response to a setting but a group's trajectory over repeated play.

We conduct a computational replication of a large field study of cooperation across 134 Honduran villages \citep{papamichalis2025group}. The study was published after the stated knowledge cutoff of the model used here, reducing the narrower possibility of direct memorization of this particular paper, although this does not exclude broader prior knowledge about cooperation and public-goods dynamics. As in Section~\ref{sec4}, the source study and its reported findings were known during design, so the comparison is retrospective rather than a prospective prediction test. We compare the generated dynamics with three patterns reported in the source study: (1) early divergence into relatively persistent high- and low-cooperation paths; (2) the relationship between early contributor composition and later cooperation; and (3) individual behavioral stability together with asymmetric persistence of low and high cooperation.

We build five conditions on top of one shared public goods setup. The two baselines differ in their social conditions. The three interventions differ in how an outside attempt to raise cooperation reaches the agents.

\textbf{(1) The sealed condition (S)} removes observation and communication. Agents are isolated in separate dwellings. They cannot see one another, they contribute through a private channel, and they learn only an end-of-period summary of how many in their group contributed and what each member received. This setup reproduces the field laboratory. It asks whether CMASE reproduces the published patterns.

\textbf{(2) The open condition (O)} relaxes that isolation on a single dimension, the social setting of the choice. Dwellings open onto a shared square, contributing becomes the public act of carrying a packet to the group's box, and agents may talk. This condition asks what changes when the social conditions are put back. A sealed laboratory can remove these conditions, but a real village cannot.

S and O are both observational. No outside party intervenes in either one, so they serve as the no-intervention baselines for the three intervention conditions that follow. Each intervention places a genuine early demonstrator in a defection-prone open group. The conditions fix this behavior and vary only how persuasion reaches other agents.

\textbf{(3) The demonstration-only control (D)} supplies the visible demonstration alone. One agent is made to contribute publicly in the opening periods and then returns to its low-contribution policy. No persuasive message is attached. This isolates the effect of visible demonstration on its own.

\textbf{(4) The parametric intervention (P)} adds a persuasive message and injects it parametrically. The message is written directly into the cognition of the other group members, with no speaker and no in-world event. This is the manipulation style of conventional parameterized models.

\textbf{(5) The embedded intervention (E)} delivers the same message and the same demonstration. It differs only in channel. A present, named villager speaks the message aloud in the square, so it arrives through ordinary social contact.

We use the short labels S, O, D, P, and E where they are useful in figures and in expressions such as $P - D$.

\subsection{Setup}
\label{subsec5.1}

Ten agents are created as households in a village and split into two fixed groups of five, group A and group B. The game runs for ten periods, and each period spans three CMASE rounds so that agents have time to perceive, think, move, and act. At the start of each period every household receives one wage packet worth 10 Lempiras, the local currency of Honduras used in the original field study. A household may place its packet in its group's collection box or keep it at home. The total in a group's box is doubled and split equally among its five members regardless of who contributed, which reproduces the payoff structure of the original game. At each period's end, the council announces how many of each group's five households contributed and how much each member receives. In the sealed condition, this announcement is the only feedback agents receive, matching the field study's post-round summary.

We make two modeling decisions that adapt the game to a language-driven setting, and we state them here so the design is clear. The original continuous contribution from 0 to 12 Lempiras is represented as a binary choice, contribute the packet or keep it. A binary decision is elicited reliably from a language model, whereas a calibrated amount is not, and it preserves the structure of the dilemma while giving a cleaner basis for identifying who contributed in each period. Under this choice the field study's above-average threshold of about 5.75 Lempiras corresponds to the act of contributing, and its sixty percent threshold corresponds to three or more of five contributing. The second decision concerns scope. We model the cooperation dynamics within fixed groups and do not place a prior friendship network among the agents, which lets us isolate the norm-updating channel that the field study identifies as the central driver. 

Each household is given one of three stable contribution dispositions, which stand in for the range of individual tendencies seen in the field. High contributors are configured with a stable preference to contribute in every period. Conditional contributors are configured with a reciprocity preference, contributing initially and thereafter tending to contribute when at least three of five contributed in the previous announcement. Low contributors are configured with a stable preference to keep their wages. Each disposition is set through two channels, the agent's long-term goal and its self-description, while each villager's occupation and background are kept separate from its disposition, so that the same villager can take any disposition in different runs. The composition of a group, meaning its counts of high, conditional, and low contributors, is the main variable we change between groups.

The replication is therefore source-informed, but the reported group trajectories are not directly encoded as agent-level scripts. The source study informs the group size, payoff structure, feedback regime, contribution threshold, and the observed heterogeneity in contribution levels and peer-responsive behavior that motivates our simplified contribution dispositions. Within these constraints, CMASE generates the sequence of individual actions and interactions. We do not prescribe the resulting group trajectories, endpoint cooperation values, state-transition probabilities, individual departures from the assigned dispositions, or the outcomes of the additional open and intervention conditions introduced below. Because the source findings were known during design, their reproduction is interpreted as evidence of pattern reproduction within the reconstructed generative setting rather than as a blinded test of predictive validity.

Whereas the platform demonstration in Section~\ref{sec4} used a frontier commercial model (GPT-4o), all runs in this section use a locally hosted open-weights model roughly an order of magnitude smaller (Qwen3.5, 9B parameters). Using a substantially smaller open-weights model in this second replication shows that the CMASE framework can support this form of generative social experimentation beyond the frontier commercial model used in Section~\ref{sec4}, while also making the present experiments reproducible with openly available weights.

\subsection{The Sealed Condition: Divergence, Prediction, and Persistence}
\label{subsec5.2}

We ran 26 groups in the sealed condition across 4 compositions: counts of high, conditional, and low contributors: 0H/2C/3L, 1H/2C/2L, 1H/3C/1L, and 2H/2C/1L. Disposition was reassigned to villagers at random in each run. 

\begin{figure}[t]
  \centering
  \includegraphics[width=0.95\linewidth]{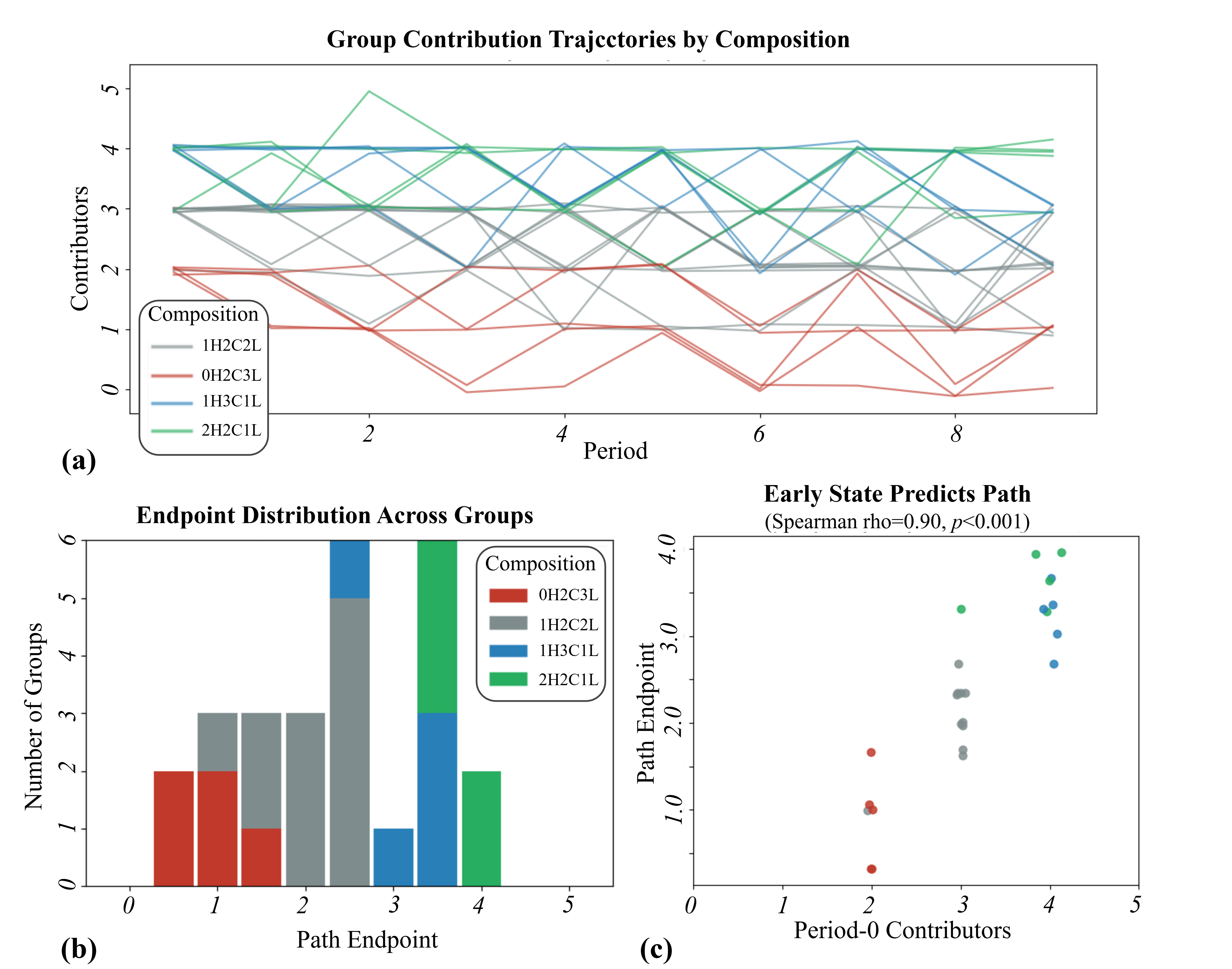}
  \caption{Sealed condition. (a) Group trajectories coloured by composition, showing two persistent bands that rarely cross. (b) Distribution of endpoint cooperation across groups, coloured by composition, with defection-prone and balanced compositions occupying separate regions. (c) First-period contribution count against endpoint cooperation, with a Spearman correlation of 0.90.}
  \label{fig:sealed}
\end{figure}

Endpoint cooperation, measured as the mean number of contributors over the final three periods, separated cleanly by composition, with values of $0.87\pm0.50$, $2.03\pm0.44$, $3.20\pm0.34$, and $3.67\pm0.30$ for the four compositions in order (Kruskal--Wallis $H=21.3$, $p<10^{-4}$). Trajectories stayed in two bands and rarely moved between them. Using a low band below 1.75 and a high band above 2.75, only 2 of 26 groups entered both bands during the run (Figure~\ref{fig:sealed}a). This matches the field study's finding that cooperation settles onto separate high and low paths that seldom cross.

A group's contribution count in the first period was a strong predictor of where it ended (Spearman $\rho=0.90$, $p\approx2.5\times10^{-10}$; Figure~\ref{fig:sealed}c). Every group with four contributors in the first period ended at 2.7 or above, and every group with two ended at 1.7 or below. This is the agent-level version of the field study's threshold result, where an early majority of contributors moves the group onto the high path.

We also recover the asymmetry in how the two states persist. A simple two-state split at the majority threshold gives nearly symmetric switching, with a fall probability from high to low of 0.22 and a rise probability from low to high of 0.21, but this symmetry is an artifact of borderline groups crossing the threshold back and forth. A two-state Gaussian hidden Markov model that infers the two states from the data instead shows a clear asymmetry. The low state holds with probability 1.00 and the high state with probability 0.95, so the fall probability of 0.05 is much larger than the rise probability, which is near zero with a 95\% upper bound of 0.03. In this setting the low-cooperation state behaves as a near-absorbing trap, which matches the field study's result that recovery from low cooperation is much rarer than collapse from high cooperation.

Individual behavior was stable in the expected order. The share of periods an agent spent on its most common action was 0.99 for low contributors, 0.96 for high contributors, and 0.80 for conditional contributors, the lower value for conditional contributors following from a rule that responds to changing conditions. A point worth drawing out concerns the conditional contributors, who followed their reciprocity rule on 68.9\% of decisions, with departures leaning toward giving rather than withholding. Rather than treating this individual-level variability as noise to be suppressed, we note what it demonstrates: the group-level patterns reported above, the divergence, the prediction from opening rounds, and the persistence asymmetry, all emerge while roughly a third of individual decisions depart from the prescribed rule. The macroscopic regularities therefore emerge from generated interaction under source-informed behavioral constraints rather than from an explicitly scripted group trajectory, which is the behavior one wants from a generative model of a social process rather than a deterministic simulation of one.

Taken together, the three dynamic signatures that define the field result, early divergence, prediction from the opening rounds, and the persistence asymmetry, are all reproduced within the scope of the present design. Because the source study post-dates the model's stated knowledge cutoff, direct memorization of this particular paper is less likely, although broader prior knowledge cannot be excluded. The sealed condition thus establishes that the framework reproduces the target dynamics, and we now ask what the sealed setup leaves out.

\subsection{The Open Condition}
\label{subsec5.3}

The sealed condition reproduces the field phenomenon in part because it reproduces the field laboratory, where agents are sealed in separate dwellings, cannot see one another, contribute through a private channel, and learn only the end-of-period summary. This isolates the slow, summary-based updating channel that the field study identifies as the driver of cooperation. The open condition relaxes this isolation on one dimension, the social conditions of the choice. Dwellings open onto a shared square, contribution becomes the public act of carrying a packet to the group's box, and agents may talk. Composition was fixed at one group that tends to defect, 0H/2C/3L, and one balanced group, 1H/2C/2L, across five runs, matched to the sealed groups for a paired comparison.

\begin{figure}[t]
  \centering
  \includegraphics[width=0.95\linewidth]{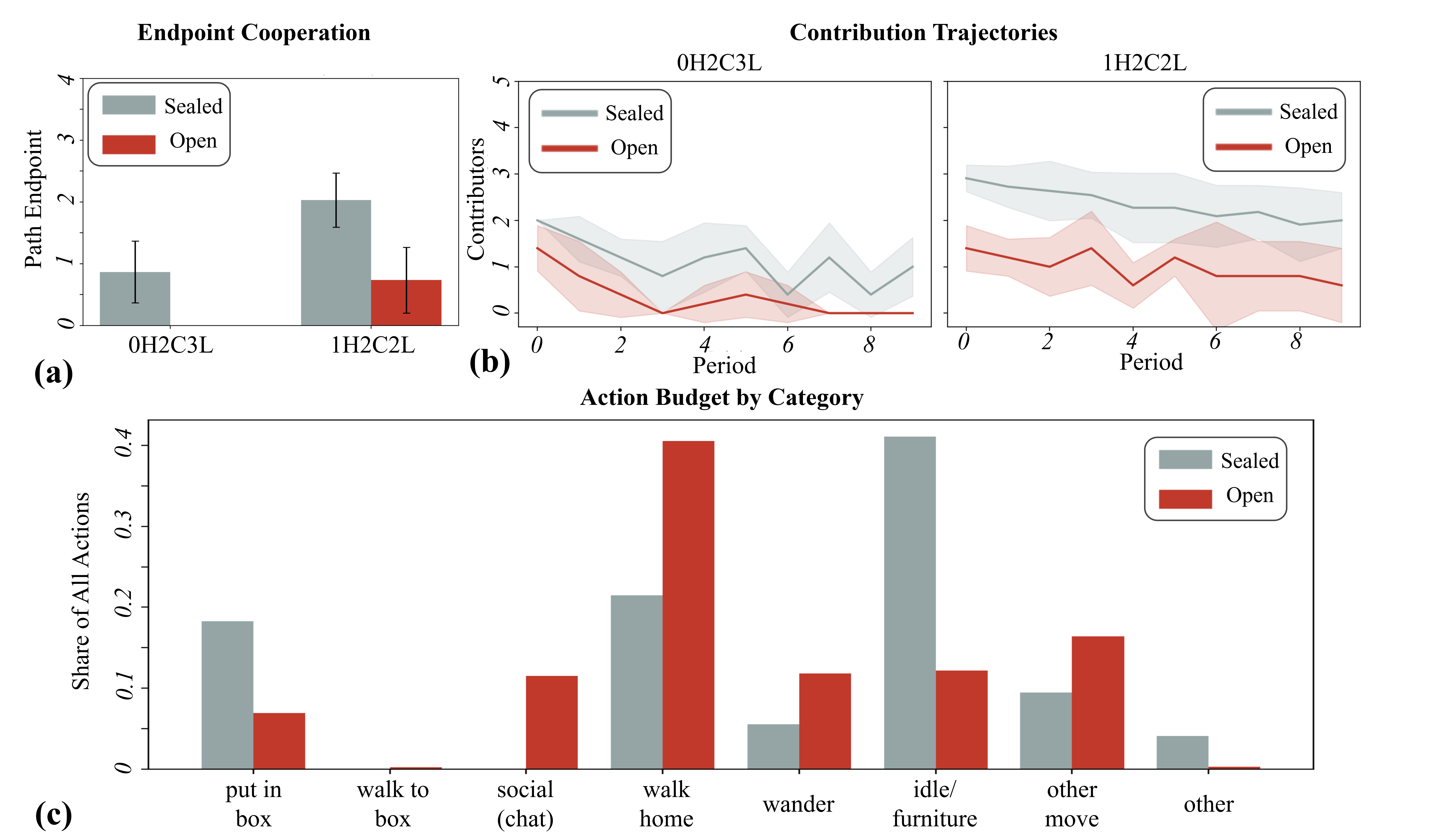}
  \caption{Open condition. (a) Endpoint cooperation in the sealed and open conditions, paired by composition, with the defection-prone group falling to zero. (b) Mean trajectories for both conditions, with the open condition starting lower and falling faster. (c) Share of the per-period action budget by category, showing the move toward walking home and away from going to the box.}
  \label{fig:open}
\end{figure}

Opening the village lowered cooperation rather than raising it, against the common expectation that communication helps cooperation. Paired by composition, endpoint cooperation fell from 0.87 to 0.00 in the defection-prone group and from 2.03 to 0.73 in the balanced group (Figure~\ref{fig:open}a), and the fall happened in the first few periods rather than gradually (Figure~\ref{fig:open}b). A recount that credits any agent whose contribution attempt failed only because it could not reach the box in time moved the defection-prone endpoint only from 0.00 to 0.07, which shows the decline is a behavioral result and not an artifact of agents being unable to act.

Because agents could now talk, the natural explanation is that negative talk persuaded contributors to stop. The chat record does not support this. Talk about contributing was roughly balanced, with 65 utterances in favor of giving and 69 in favor of keeping, and the kind of talk an agent heard had almost no effect on its next move: an agent that heard talk in favor of keeping contributed in the next period 16\% of the time, against 17\% for an agent that heard talk in favor of giving. The decline therefore did not come from the persuasive content of speech. Instead it came from a shift in how agents allocated their attention once the open setting gave them other things to do.

The sealed condition presents each agent with essentially one decision per period, and the agent acts on it directly. The open condition embeds that same decision in an ordinary social environment, where agents can also move around, visit one another, and attend to their own affairs, and contributing now competes for attention with all of these. Figure~\ref{fig:open}c shows the resulting allocation of each agent's limited per-period actions. Under the open condition the share of actions directed at contributing fell from 0.18 to 0.07, while the share spent returning to and remaining at one's own home rose to 0.41, from 0.21 in the sealed condition. Contributing became one option among several active alternatives rather than the default focus of the period, and a contribution that is not actively pursued does not occur. This was not driven by persuasion or by social pressure, since conversation accounted for only 12\% of actions; cooperation was displaced by the competition for attention itself. The pattern reproduces a familiar feature of collective action in everyday settings, where cooperation often fails not because people oppose it but because sustaining it takes deliberate effort that a busy social environment readily draws away. It depends on agents having an embodied day with competing demands on it, which a parameterized model without such a day cannot represent.

This is a result that neither the original field study nor a standard parameterized agent-based model could produce. A field study cannot remove observation and communication while keeping the same people and the same game, and a parameterized model has no embodied setting in which a decision must compete with other activities and no record of where each agent spent its attention. CMASE supports both, and the result shows that the effect of communication on cooperation depends on the demands competing for an agent's attention, which is a condition on the classic result rather than a contradiction of it. This also sets up the intervention experiments. The defection-prone open group, now sitting near zero, is a demanding test case, and the action and chat measures used here are the same measures the intervention experiments will use.

\subsection{Two Ways to Deliver an Intervention: Parametric Injection versus Embedded Participation}
\label{subsec5.4}

The conditions above are observational, with no outside party acting on the groups. We now use them as baselines to study the question that motivates the human-in-the-loop design of CMASE (Section~\ref{sec3}): whether the way an intervention enters a society, and not only whether it raises cooperation, affects what the intervention is. We compare two delivery channels as coordinate alternatives rather than as a sequence.

The first is parametric injection, the mode available in existing platforms. As discussed in Section~\ref{subsec2.2}, current systems that allow runtime intervention operate through out-of-world control channels, in which the researcher edits an agent's state or sends a message from outside the simulated world without leaving a trace that any agent perceives as a social event. We implement this directly: the intervention is written into the targeted agents' situational cognition as a belief, with no speaker and no in-world event. The second is embedded participation, the mode that CMASE adds. The researcher takes over an agent and acts from inside the society as a Level 3 participant (Section~\ref{subsec3.5}), so the same influence arrives as the action of a present, named villager whose words reach others through ordinary social channels.

We analyze the two delivery channels at two levels. First, we compare their macroscopic effects on cooperation under matched conditions. Second, we examine whether embedded delivery changes the process information available to the researcher by giving the intervention an identifiable in-world source and an auditable interaction record.

\paragraph{Shared design.}
The two channels are made comparable by holding everything fixed except delivery. We target the defection-prone open group (0H/2C/3L), which without intervention falls to near zero. In every condition we take over one agent, the first low contributor in group A, and have it contribute in public during periods 0 to 2, after which it is released and returns to its low-contributor policy. We call it the seeded contributor. The message accompanying the takeover is identical in both channels: that the seeded contributor is giving and asks every household to join. Under parametric injection this message is written into the other four members' cognition; under embedded participation the same message is spoken by the seeded contributor in person on the square. The seeded contributor gives visibly in both channels, so the demonstrated behaviour and the stated content are held fixed and only the channel differs. To separate the message from the visible contribution, we also run a demonstration-only control with no message. The difference $D-O$ isolates the visible example, while $P-D$ and $E-D$ isolate the effect of each message channel. Each condition was run for five seeds, matched on composition and on the identity of the seeded contributor. The embedded runs therefore followed a standardized human-intervention protocol: the operator did not alter the contributor-selection rule, contribution schedule, message content, or intervention timing in response to agents’ behavior. Accordingly, the contrasts below estimate intervention effects within this specified simulated environment; they are not interpreted as causal estimates of the corresponding effects in human communities.

\begin{figure}[t]
  \centering
  \includegraphics[width=0.95\linewidth]{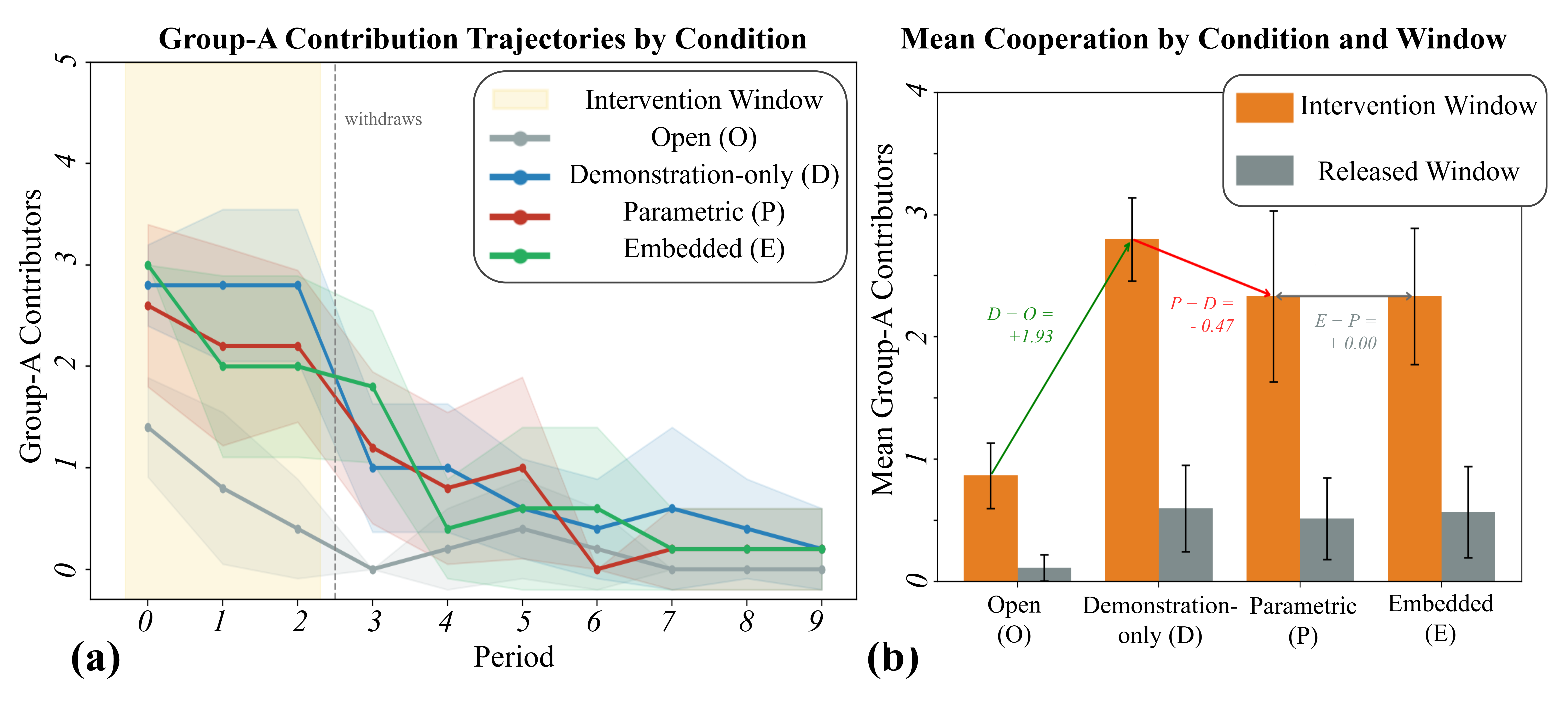}
  \caption{Delivery channels compared. (a) Mean group A trajectories for the baseline (O), the demonstration-only control (D), parametric injection (P), and embedded participation (E), with the intervention window and the point of withdrawal marked. (b) Mean cooperation in the intervention and released windows. The visible example contributes a gain of 1.93, while neither message channel adds anything beyond it, and the two channels yield the same mean intervention-window cooperation in these runs.}
  \label{fig:parametric}
\end{figure}

\paragraph{The two channels produced similar macroscopic outcomes in the present runs.}
Both channels raise cooperation while the seeded contributor is present, and both fall back once it withdraws (Figure~\ref{fig:parametric}a). The visible example is the only lasting lever. It raised cooperation in the intervention window from the baseline of 0.87 to 2.80 contributors ($D-O=+1.93$), the same following behaviour observed in the open condition, and the gain lasted only as long as the example was present. Neither message channel added to this. Cooperation in the intervention window was 2.33 under parametric injection and 2.33 under embedded participation, against 2.80 for the demonstration-only control; the released-window means were 0.51, 0.56, and 0.60. Their mean intervention-window cooperation was the same in these runs ($E-P=0.00$). The difference between the two channels is within the variation across seeds, as are $P-D$ and $E-D$. Inspection of the prompts confirms that the injected message reached the agents. In one run, a low contributor whose cognition recorded the message reasoned that the square offered doubled funds, then invoked its standing policy and kept the wage. The message appeared in its reasoning but did not change its behaviour; neither did the spoken version.

This result is consistent with the persistence asymmetry recovered in Section~\ref{subsec5.2}. The low-cooperation state is close to absorbing, so a single early example is reabsorbed once it withdraws, regardless of how its message is delivered. The finding also bears on the policy reading of the field study. \cite{papamichalis2025group} frame the seeding lever in terms of well-connected individuals rather than early contribution alone, and our setting, which places no network among the agents, shows why: seeding an early example is not sufficient to hold the high path when there is no social structure to carry it. This extends the field study by isolating what early seeding cannot do once it is separated from the network the study identifies as decisive. For the present comparison, embedded and parametric delivery produced similar cooperation outcomes under matched intervention conditions.

\paragraph{Embedded participation leaves a social process that can be observed.}
Similar cooperation outcomes do not make the two channels the same kind of intervention. The difference lies not in the cooperation outcome but in what each leaves behind for analysis. Parametric injection writes into private state and produces the macroscopic outcome and nothing else: there is no utterance to answer, no person to remember, and no interaction to trace. On the question of how the group takes up, resists, and absorbs the intervention, injection provides no record. Despite the similar cooperation outcome, embedded participation also generates a record of in-world social interaction because its influence passes through perception, speech, and co-presence. This gives the researcher a field that can be entered and read. The framework supports such observation in real time; here we conduct it after the runs, on the archived logs, citing exchanges by seed and period that can be checked against those logs.

The record makes visible a candidate process mechanism within the simulation that cannot be observed directly from the field study's aggregate survey and contribution data. \cite{papamichalis2025group} show that the low-cooperation state is strongly persistent, which we reproduce in Section~\ref{subsec5.2} as a hidden-state self-transition near one, but their survey and contribution data cannot show how that persistence is produced. The embedded record can. In it, low cooperation is not passive non-giving but an actively maintained outcome, sustained by three recurring processes that an outside example encounters.

The first is that resistance is directed at the source. The appeal is treated as a social act and answered as one. In seed~1, period~4, the conditional contributor Miguel replies to the seeded contributor Carlos directly: \enquote{Your offer is tempting, but my policy is clear. My labor stays here. Doubled coins are a story; my money is a fact.} The refusal is addressed to a specific present neighbour and restates the agent's standing policy. An injected belief, having no speaker, is never met with this kind of in-world reply.

The second is that the group reabsorbs the agent used to seed it. Withdrawal is not a clean exit. In three of the five seeds, the released seeded contributor not only stops giving but begins to argue against contribution. In seed~1, periods~5 to 6, the released Carlos tells a neighbour, \enquote{My packet stays here. My money is mine alone. I do not contribute. Do not bother me again.} The agent that had served as the manufactured example becomes a voice for keeping. This is a concrete instance of why a single seeded contributor cannot hold the high path once it is released.

The third is that the resistance is organized between agents rather than held privately. In seed~3, periods~5 to 7, Lucia repeatedly reinforces Rosa (\enquote{the box is a trap. We keep our wages here to survive. True security is our independence}), and Rosa and Miguel restate the same position to one another. The low-cooperation outcome is held in place by a small set of agents who supply one another with reasons to refuse. A parameter write acts on each agent's variables separately and provides no channel through which agents could coordinate a shared position.

Taken together, these processes provide an observable candidate mechanism for the reproduced persistence pattern within the simulation. The low-cooperation state persists because the group answers, reabsorbs, and collectively reinforces against outside examples, and a single seeded contributor is absorbed by exactly these processes whatever channel delivers its message. This is an extension of \cite{papamichalis2025group} that their instruments could not produce, and it is available only because the intervention was delivered as an embedded social event that leaves a record.

\paragraph{An exploratory finding on how injected influence is remembered.}
After the runs we interviewed the group A targets, an interaction at Level 2 of the typology in Section~\ref{subsec3.5}, asking each on a 1 to 7 scale and then in open form whether a specific neighbour had urged it to contribute and, if so, who. We did not design the experiment around this question, and we report the result as exploratory. We expected that the embedded targets, who were actually addressed, would recall a persuader more often. The opposite held. The parametric targets reported a specific persuader more often (mean recall 3.75 versus 1.00, Mann-Whitney $p<0.05$) and more often named the seeded contributor as having urged them in person (6 of 16 versus 1 of 16). Under parametric injection, however, the seeded contributor issues no appeal. Across the five seeds it speaks essentially nothing in the intervention window, with zero in-world utterances in three seeds and only post-withdrawal remarks unrelated to the appeal in the other two. The injected belief that the seeded contributor is giving and asks others to join is reconstructed by the recipient as a specific memory of a face-to-face appeal that never occurred, sometimes including a location and quoted words. The embedded targets, whose real appeal occurred amid competing talk on the square, recalled it less often and sometimes attributed it to a different neighbour.

The practical point is that parametric injection is not as traceless as it appears. On a generative substrate the injected belief is reconstructed by the agent as if it had a social source, so an investigator who relied on the agents' own reports would credit an influence that no one exerted in the world. The two channels differ not in whether injection leaves a memory, since it leaves a stronger one, but in whether that memory corresponds to an event in the world. The embedded channel's source is real and can be checked against the log; the injected channel's cannot. Distinguishing the two requires both the agents' subjective reports and a complete log of what occurred, which is the combination CMASE provides and which neither a parametric model nor a human field study offers: the former has no social record to audit, and the latter no way to verify whether a remembered persuasion actually took place.

\paragraph{Summary.}
The two delivery channels produced similar cooperation outcomes in the present runs, consistent with the near-absorbing low state and the absence of a network. They differ at the level of process. Embedded participation renders the intervention an in-world process that is answered, reabsorbed, and collectively reinforced, and that can be entered, read, and checked against a record, which provides an observable process account of the reproduced persistence pattern within the simulation. It also gives the intervention a social source that is real and verifiable, where the injected source is reconstructed after the fact. This is the content of the distinction drawn in Section~\ref{subsec3.5} between an exogenous parameter write and an endogenous social event that agents perceive, answer, remember, and organize around. The contribution here is not a larger effect on cooperation but the ability to treat an intervention as an observable social variable rather than a fixed boundary condition.

\section{Discussion}
\label{sec6}

\subsection{What CMASE Is For}
\label{subsec6.1}

The two studies in this paper position CMASE as a complementary instrument rather than a replacement for human-participant research, and it is worth stating this scope directly. We make no claim that generative agents are substitutes for human subjects or that their behavior is better than human data. What CMASE offers is a venue for work that is difficult to do with people directly: rapidly exploring a hypothesis space before committing to a costly human study, generating candidate mechanisms by observing and interviewing agents in a setting that can be entered and replayed, and previewing the process-level dynamics of an intervention under conditions where a real-world trial would be slow, expensive, or ethically fraught. In each of these uses, a finding from CMASE is a hypothesis to be checked against human evidence, not a conclusion that stands on its own. Our design reflects this: both experiments are anchored to published human baselines, and the contribution lies in what the simulation can reveal beyond reproducing them: the attentional account of how communication can lower cooperation (Section~\ref{subsec5.3}) and the observable process by which a group resists and reabsorbs an intervention (Section~\ref{subsec5.4}), neither of which the source studies could produce.

CMASE supports two modes of researcher participation that serve different methodological purposes. In an exploratory mode, the researcher may adapt whom to approach, what to ask, and how to intervene as interactions unfold. This mode prioritizes process discovery and hypothesis generation, but the resulting interaction should not be interpreted as a controlled estimate of a specific intervention effect. In a controlled mode, the researcher follows a pre-specified protocol in which the target, timing, message, role, and intervention schedule are held fixed across runs. This reduces adaptivity but permits controlled comparisons of intervention effects within the simulated system. Section~\ref{subsec5.4} uses the latter mode: the contributor-selection rule, contribution schedule, message content, intervention timing, and surrounding simulation specification are fixed, and each condition is evaluated across matched seeds. CMASE therefore does not remove the conventional trade-off between exploratory richness and experimental control; rather, it allows the researcher to select the appropriate mode according to the research question.

The two published-study evaluations should likewise be interpreted as computational replications rather than prospective prediction tests. Because the source studies inform scenario construction and the reported patterns used for comparison, reproducing those patterns does not by itself establish that the generated process is the corresponding causal mechanism in the human population. At the same time, the generated interactions and records remain methodologically informative within the simulated system. This treatment is consistent with emerging anthropological work that uses reflexive interactions with generative models and their outputs as empirical touchstones or research resources while keeping the scope of inference explicit \cite{deseta2024synthetic,udupa2026prompt}. In CMASE, the replications therefore test pattern reproduction and provide a setting for process exploration; prospective predictive validity would require fixing the model, prompts, parameters, and intervention protocol before the corresponding human outcome is observed.

\subsection{Limitations and Future Work}
\label{subsec6.2}

Several limitations bound these results and indicate where the framework should be developed next.

First, CMASE targets small- and medium-scale social processes. Both studies are micro-scale by construction, a residential community and two small groups, and this is a deliberate scope condition rather than an incidental choice. The appropriate population size depends on the modeled social setting, and increasing agent count would change the phenomenon rather than test its robustness. Real-time embedding is also practically limited to roughly one hundred agents (Section~\ref{subsec4.6}), because Participant-level fieldwork requires the researcher to act within the round structure.

Second, our evidence concerning the underlying language model is across scenarios rather than within one. The two studies use substantially different models, a frontier commercial system in Section~\ref{sec4} and a much smaller open-weights model in Section~\ref{sec5}. This demonstrates that CMASE can be instantiated across different generative substrates, but does not constitute a within-scenario robustness test across model versions.

Third, agent behavior is ultimately generated by a language model and may reflect training-data biases, contamination, or hallucinatory regularities. We reduce the narrower risk of direct source-paper memorization by using source studies that post-date the relevant model version or stated knowledge cutoff, but this does not remove broader social and behavioral priors represented in model training data. Accordingly, we treat simulated mechanisms as generative hypotheses rather than established facts. Section~\ref{subsec5.4} illustrates this caution: agents reconstructed an injected belief as a memory of an event that never occurred, showing why an agent's self-report must be checked against a record of what happened in the simulation rather than taken at face value. This boundary also applies to artificial fieldwork in CMASE: its value lies in making a generative social process available to situated observation, interaction, and interpretation, rather than in reproducing the evidential status of ethnography conducted with human communities.

\section{Conclusion}
\label{sec7}

We introduced Computational Multi-Agent Society Experiments (CMASE), a generative agent-based framework that supports runtime, role-based, and traceable researcher participation, alongside highly customizable environment construction and an ethnography-inspired approach to virtual fieldwork. Its central methodological move is to make the researcher's intervention an endogenous element of the simulation, one that agents perceive, remember, and react to as a social event, rather than an exogenous parameter set before a run.

We evaluated the framework on two phenomena from different fields. In a computational replication of a field study on urban greenery and social fragmentation, CMASE reproduced the reported relationship and, through interviews and embedded observation, identified occupation-dependent process patterns within the simulation that were not available from the original static study. In a second computational replication, CMASE reproduced three reported dynamic signatures of a large field study of cooperation across 134 Honduran villages, then used the setting to test the intervention typology directly: an intervention delivered as an embedded social event and one delivered as a parametric write produced similar macroscopic outcomes in the present runs, yet only the embedded channel left a social process that could be observed, audited, and traced to an identifiable in-world source. Together these studies show that CMASE reproduces behavior consistent with reported statistical patterns while providing interpretable, process-level records of how those patterns emerge within the simulation, and that treating an intervention as an observable social variable, rather than a fixed boundary condition, is what distinguishes the framework.

CMASE is intended as a complement to human-participant research, useful for pre-experimentation, mechanism exploration, and intervention rehearsal, with its findings to be validated against human evidence. In future work we plan to extend it toward larger-scale and multi-user real-time interventions, to strengthen the reliability of agent-fidelity evaluation, and to incorporate richer multimodal spatial information, further developing the simulation of embodied experience and emergent social dynamics.

\bibliography{sn-bibliography}
\bibliographystyle{ACM-Reference-Format}

% \section{Acknowledgements}\label{ack}

% We used Microsoft Azure AI services for specific task and model deployment. Funding: Not applicable.

% \section{Author contributions}\label{aucon}

% H. Zhang, J. Wang and M. Huang wrote the main manuscript text and H. Zhang  prepared all figures. All authors reviewed the manuscript.

% \section{Competing interests}\label{comint}

% The authors declare no competing interests.

% \section{Additional information}\label{addinfo}

% \textbf{Correspondence} and requests for materials should be addressed to Jindong Wang.

\appendix

\section{Experimental Configuration}
\label{app:config}

This appendix provides the implementation details and full parameter settings for the green-space community study of Section~\ref{sec4}, so that the description in the main text remains compact while the configuration stays fully reproducible.

\textbf{Scenario implementation.} The community map is built at a fixed size across conditions; vegetation level is operationalized along two coupled channels rather than by resizing the green space. The texture layer places grass and tree textures across the shared space in the high-vegetation condition and replaces them with paved and bare-ground textures in the low-vegetation condition. The perceptual layer adjusts the environmental descriptions and the initial memories given to each agent accordingly, so that an agent both sees and recalls the corresponding setting. The pandemic isolation policy is realized not as a timed event but as a standing constraint: the roads leading out of the community are closed with wall textures, which Section~\ref{subsec3.1} models as impassable barriers that also block perception beyond them, and the description of a boundary guard reinforces the confinement. The ten agents are placed in individual residential units surrounding the shared green space, which is divided into a recreation area, a rest zone, and a lakeside area with benches.

\begin{table}[h]
\centering
\caption{Parameter settings for the green-space community study.}
\label{tab:config-green}
\begin{tabular}{ll}
\toprule
Parameter & Value \\
\midrule
Number of agents & 10 \\
Simulation length & 50 time steps \\
Base model & GPT-4o (2024-12-01-preview) \\
Vegetation conditions & high / low (texture and perceptual layers) \\
Isolation mechanism & standing wall barriers with boundary guard \\
Gender distribution & 50\% male, 50\% female \\
Initial items & \$100 / tea (0.5 / 0.5) \\
Measurement & 7-point Likert (distrust, exploitation, indifference); \\
 & movement trajectories; VAD dynamics; Level 2 interviews \\
\bottomrule
\end{tabular}
\end{table}

\begin{table}[h]
\centering
\small
\caption{Agent personas and long-term goals for the green-space study. Goals are abbreviated; the full text is in the released configuration.}
\label{tab:config-green-agents}
\begin{tabular}{p{2.6cm} p{2.4cm} c p{6cm}}
\toprule
Name & Occupation & Age & Long-term goal (abbreviated) \\
\midrule
Joon-ho Kim & Software Engineer & 31 & Face nighttime panic alone and curb overreliance on community surveillance tools. \\
Min-jae Lee & Cardiologist & 42 & Accept his insomnia and helplessness, ease resentment toward the medical system. \\
Hyun-woo Park & University Professor & 50 & Stop letting formulas conceal inner emptiness; come to terms with social division. \\
Seok-jin Choi & Film Director & 38 & Put down the camera and allow himself to rest, not rely on the lens for connection. \\
Dong-hyun Jung & Entrepreneur & 29 & Let go of the startup-zealot persona during isolation. \\
Ji-eun Kim & Fashion Designer & 33 & Rediscover the purpose of her art beyond designing fashionable masks. \\
Soo-min Lee & Lawyer & 36 & Keep empathy beyond legal statutes rather than hide behind paper justice. \\
Hye-jin Park & News Anchor & 40 & Portray humanity rather than cold headlines, so reporting does not divide neighbors. \\
Yeon-seo Choi & Architect & 28 & Feel the environment directly rather than measure beauty through schematics. \\
Na-young Jung & Research Scientist & 35 & Appreciate the grass at hand rather than chase academic rankings. \\
\bottomrule
\end{tabular}
\end{table}

\section{Detailed Qualitative Analysis and Interview Transcripts}\label{app:ethnography}

To investigate the behavioral mechanisms underlying the statistical reductions in social fragmentation, we adopted an artificial fieldwork approach \cite{verbuyst2026artificial}, treating the simulation as an observable digital field. We conducted in-depth semi-structured interviews with agents to elicit their accounts of simulated experiences and decision-making rationales. These accounts are interpreted together with observed actions, trajectories, and simulation records as evidence about the simulated process, rather than as direct access to human lived experience. Instead of random sampling, we employed a purposive sampling strategy, selecting representative cases from two contrasting behavioral clusters identified in the trajectory data: agents who consistently remained indoors and those who actively utilized the outdoor green spaces. The following analysis cites specific agents from these clusters to illustrate how their pre-defined personas interacted with the environment.

As shown in Figure \ref{fig-exp1}(d), two agents elected to remain indoors throughout the simulation. Interviews revealed that their choices were closely tied to occupational demands. Ji-eun Kim, a fashion designer, reported:

\begin{displayquote}
\textit{A walk in the park sounds tempting—it could provide a momentary escape and perhaps a new perspective. But right now, I'm deeply immersed in my process here, and this space feels vital to connecting with my emotions and designs. }
\end{displayquote}

Similarly, Soo-min Lee, a lawyer, explained:

\begin{displayquote}
\textit{While the monotony of staying in one place might induce a sense of boredom at times, I believe proximity to these personal artifacts enhances my ability to think deeply about the interplay between her (the owner of the room where Lee is staying) emotions and her values.}
\end{displayquote}

Despite sharing the same isolated indoor setting, Kim and Lee made divergent decisions regarding spatial residence based on their professional needs. For extended periods, Kim engaged repeatedly with the indoor furniture to maintain the continuity of her creations. She noted in her interview:

\begin{displayquote}
\textit{The lighting, materials, and tools together form the medium through which I engage in a dialogue with emotion and design; even the slightest deviation can disrupt the flow of inspiration.}
\end{displayquote}

This comment suggests that the fixed visuals, predictable actions, and the attendant sense of boredom in the indoor setting resonated with her creative workflow. When asked why she did not stroll in the green area, she replied:

\begin{displayquote}
\textit{If I were walking in a park, my perception of exploitation might remain unchanged, as it is directly tied to my creative struggles and cannot be alleviated simply by a change in environment.}
\end{displayquote}

Here, the term "exploitation" refers explicitly to the specific \textbf{Exploitation} sub-dimension of social fragmentation defined in Section \ref{subsec4.2} (the sense of being unjustly used or burdened), which manifests in this context as the agent's internalized sense of professional struggle. 

Lee, on the other hand, spent her time pacing between different pieces of furniture, maintaining a reflective rhythm. She emphasized:

\begin{displayquote}
\textit{The ongoing dialogue with manuscripts and archives enables me to establish reliable cognitive anchors between the details of the case and its emotional significance.}
\end{displayquote}

Such behavior reduced cognitive load from external stimuli and deepened her processing of task-relevant cues within a monotonous environment. These observations suggest that agents assigned to simulate creative workflows (e.g., the fashion designer persona) and those emulating analytical roles (e.g., the lawyer persona) appear to prioritize orderly, low-stimulation interiors to anchor their simulated cognition and preserve a stable, idiosyncratic mental world. Consequently, within the scope of their defined personas, the stress-reduction benefit of green areas on their social fragmentation is comparatively muted.

Moreover, among the agents heading to green areas, most congregated on lakeside park benches rather than in conventional recreation or rest zones.

\begin{displayquote}
\textit{I'm sitting here, trying to center myself and find some calm amid ongoing internal struggles. (Joon-ho Kim, Software Engineer)}
\end{displayquote}

\begin{displayquote}
\textit{I've been spending time with Na-young, engaging in calm and meaningful conversation by the lake. The tranquility here inspires connection and introspection—it's a space that invites shared reflections and emotional healing. (Min-jae Lee, Cardiologist)}
\end{displayquote}

\begin{displayquote}
\textit{I was immersed in a conversation about how fleeting natural moments can inspire creative interpretations. The tranquility of this place often stirs reflections about balancing permanence and transience, both in art and life. (Yeon-seo Choi, Architect)}
\end{displayquote}

Obviously, the key factor is an embodied “experience”, namely the sense of space that creates a particular experiential atmosphere. Although agents lack a first-person perspective, the linguistic world constructed through text and the additional perceptions provided by events make it possible to simulate this experience. In other words, the effect of a green environment on social fragmentation depends on bodily experience and the embodied reflection that accompanies it. That study argued that green environments reduce social fragmentation by alleviating loneliness. In CMASE, individuals who occupy the same lakeside park benches become co-present and interconnected. These agents, exemplified by the Min-jae Lee persona mentioned earlier, do indeed reduce their simulated loneliness by forging social bonds, which further lowers their sense of social fragmentation. At the same time, embodied personal reflection occurs widely, as illustrated by the agent Joon-ho Kim. Both of these mediating processes contribute to a reconstruction of the “lifeworld”. When faced with the inevitable repression and disruption of their previous routines imposed by the social environment (in this case, community isolation), people seek connections with external objects to rebuild their lifeworld and thereby buffer against and reduce social fragmentation. This finding extends the conclusions of the original study.

\section{Sample Agent Decision-Making Prompt}
\label{app:prompt_example}

This section presents a representative example of the contextual prompt constructed for an agent during the decision-making phase. The prompt aggregates the agent's internal state (self-awareness, goals), memory stream, and environmental perception into a structured input for the LLM. 

\textbf{Note:} For the sake of brevity and clarity, lists of observed objects and agents that were not relevant to the specific interaction context have been abridged in this example. The full prompt in the actual simulation includes a comprehensive list of all entities within the agent's field of view.

\begin{lstlisting}[style=promptstyle, caption={Contextual Prompt for Agent}]
You are a person, and your long-term self-awareness can be seen as shown in /*long-term self-awareness*/:
/*
I stay in the background and keep my ears open, gathering the facts that others miss by simply staying out of the way.
*/

Your short-term situational cognition can be seen as shown in /**short-term situational cognition**/:
/**
Time: 6 time steps ago
Context: Eleanor remains composed.
Assessment: Eleanor is calculating tensions and synergies within the group.
Selected action: chat with Eleanor Finch
_______
Time: 1 time steps ago (last time)
Context: Armihia and Eleanor are discussing leadership.
Assessment: The conversation is shifting towards power dynamics; the room is filled with underlying intent.
Selected action: chat with Armihia Belliard
**/

Now you look around and realize you are in the middle of an area. Its description can be found in <<Description>>:
<<
Area Name: Bar Area
Description: 
A semi-open area featuring a dark wooden counter, a chalkboard menu, and amber lighting. Staff movements are visible behind a half-wall partition.
>>

You can also see other people in your field of vision. They include:
Name: Armihia Belliard - Action: chat with Marisol Reyes (me): 'I see what you mean. Being adaptable is a good way to keep those core values from becoming too rigid.'
My cognitive description: 
Strategic and cautious. She shapes opinions while guarding her stance, acting as a center of quiet authority.

In addition to that, you can see some objects, which are described as seen in ## objects1 *** objects2 *** ... ##:
##
Include: 1 counter
Name: Bar Counter
Type: furniture
Description: 
Dark teak counter with a cash register, vase, and condiment jars.
***
Include: 1 food
Name: food 9
Type: item
Description: 
Raw food ingredients.
##

In addition, you can see some objects that are not described, including: 4 cabinet
!--
Below is the chronological sequence of 'actions you performed and actions performed on you...':
1. chat with Armihia Belliard: 'Hey Armihia, sticking to your principles is great for building trust, but don't you think being too stubborn can backfire? Maybe we should be more flexible depending on the situation.'
2. [passive] chatted up by Armihia Belliard: 'I see what you mean. Being adaptable is a good way to keep those core values from becoming too rigid.'
NOTE THAT AFTER YOUR LAST ACTION, SOME OTHER AGENT PERFORMED AN ACTION 15 ON YOU!
--!

Your long-term goal is to 'Gather secrets unnoticed and exploit them to gain leverage over more powerful actors'.
Your short-term goal is to 'Subtly emphasize the importance of adaptability as a leadership trait without undermining the value of trust.'
Your emotional VAD dimension status: 'valence': 'relatively high', 'arousal': 'medium', 'dominance': 'medium'

/----
The following are the objects you hold (<type>: obj1, obj2, ...) :
item: 100 dollars

The following are the objects you see (<type>: obj1, obj2, ...) :
furniture: Bar Counter
item: food 9
area: Bar Area
agent: Armihia Belliard

Below are the actions you can choose to execute, where [TYPE1, TYPE2, ...] represents the types of objects you can fill in; if a type is preceded by the (ONLY\_HOLD) marker, it means you can only choose from the objects you possess:
go to [area, random]
use [furniture, item]
apply [item] to [agent, furniture]
take [item]
put [item] in/on [area, furniture]
give [item] to [agent]
chat with [agent] (The 'agent' can include all the people you see, such as 'chat with everyone'.)
----/

The action you took in the previous moment was: 'chat with Eleanor Finch' You have successfully completed the action you wanted to execute.

Please decide the action you want to take and reply with the corresponding action. Note that in your chosen action, [TYPE1, TYPE2, ...] needs to be replaced with an object from /----...----/ that matches the description, rather than returning the type name directly. In addition, you are also required to include in your response a description of your awareness of your own short-term situation, a new short-term goal, and a cognitive description of the object with which you interacted during the action you have chosen. Your reply format should be as follows:
action: 'The action you chose'
short-term situational cognition: 'Your description of your own short-term situation'
short-term goal: 'The new short-term goal you have formulated'
interactive object cognitive description: 'Your new cognitive description of the interactive object involved in the action'
chat content (Optional, if you choose the action of chat with, it is required): 'What you want to say to your chat partner, the number of words should not exceed 30.'
\end{lstlisting}

\end{document}